%% file: main.tex
\documentclass[11pt]{article}

\usepackage[T1]{fontenc}
\usepackage{lmodern}

\usepackage[margin=1in]{geometry}
\usepackage{microtype}
\usepackage{indentfirst}

\usepackage{amsmath,amssymb}

\usepackage{booktabs}
\usepackage{tabularx}
\usepackage{longtable}
\usepackage{pdflscape}
\usepackage{array}
\usepackage{enumitem}

\usepackage{xcolor}
\usepackage{titlesec}

\makeatletter
\long\def\@makecaption#1#2{%
  \vskip\abovecaptionskip
  \sbox\@tempboxa{{\small\bfseries #1} {\small #2}}%
  \ifdim \wd\@tempboxa >\hsize
    {\small\bfseries #1} {\small #2}\par
  \else
    \global \@minipagefalse
    \hb@xt@\hsize{\hfil\box\@tempboxa\hfil}%
  \fi
  \vskip\belowcaptionskip}
\long\def\LT@makecaption#1#2#3{%
  \LT@mcol\LT@cols c{\hbox to\z@{\hss
    \parbox[t]\LTcapwidth{%
      \normalfont\small
      \sbox\@tempboxa{{\bfseries#1{#2: }}#3}%
      \ifdim \wd\@tempboxa >\hsize
        {\bfseries#1{#2: }}#3\par
      \else
        \hbox to\hsize{\hfil\box\@tempboxa\hfil}%
      \fi
      \endgraf\vskip\baselineskip}%
    \hss}}}
\makeatother

\usepackage[hidelinks]{hyperref}
\hypersetup{pdfauthor={Anonymous Authors},
            pdftitle={Retrieval-Augmented Generation: A Review in the Agentic Era}}

\usepackage[numbers,sort&compress]{natbib}

\usepackage{graphicx}
\usepackage{tikz}
\usetikzlibrary{arrows.meta,positioning,shapes.geometric,fit,backgrounds,calc}
\usepackage{adjustbox}

\definecolor{BurntOrange}{RGB}{204,102,0}

\usepackage{pifont}

\titleformat{\section}{\normalfont\Large\bfseries}{\thesection}{0.6em}{}
\titleformat{\subsection}{\normalfont\large\bfseries}{\thesubsection}{0.5em}{}
\titleformat{\subsubsection}{\normalfont\normalsize\bfseries}{\thesubsubsection}{0.5em}{}

\newcommand{\cmark}{\ding{51}}   
\newcommand{\xmark}{\ding{55}}   
\newcommand{\pmark}{$\triangle$} 

\begin{document}

\begin{center}
{\LARGE\bfseries Retrieval-Augmented Generation in the Agentic Era:\\[4pt]
Tradition, Present, and Prospect\par}
\vspace{26pt}
{\large
\textbf{Chenming Tang}$^{1}$\quad \textbf{Haolin Li}$^{1}$\quad \textbf{Yiran Li}$^{1}$\quad
\textbf{Zixian Yang}$^{1}$\\
\textbf{Lizhu Zhai}$^{2}$\quad \textbf{Rui Sun}$^{2}$\quad \textbf{Qizhou Huang}$^{2}$\quad \textbf{Yunfang Wu}$^{1,\dagger}$\par
\vspace{5pt}
{\small
$^{1}$\textit{State Key Laboratory of Multimedia Information Processing, School of Computer Science, Peking University}\quad $^{2}$\textit{China United Network Communications Group Co., Ltd.}\par}
\vspace{5pt}
{\small
\texttt{tangchenming@stu.pku.edu.cn\quad wuyf@pku.edu.cn}
}
}
\end{center}
\makeatletter
{\def\@thefnmark{}%
 \insert\footins{\normalfont\footnotesize
   \@makefntext{
   $^{\dagger}$Corresponding author.}}}
\makeatother
\vspace{10pt}

\begin{center}
\begin{minipage}{0.86\textwidth}
{\bfseries Abstract.}\quad
Retrieval-augmented generation (RAG) couples frozen language models with external, updatable knowledge, so that generation is grounded in retrieved evidence rather than in parametric memory alone.
This work maps RAG in the agentic era by joining them on a single lifecycle axis, from offline indexing through pre-retrieval, retrieval, post-retrieval, and orchestration over these stages. It treats indexing and orchestration as first-class components and agentic workflows as a paradigm of their own, and it covers the agentic work that has appeared since those reviews. Beyond these components, it covers evaluation, applications, and the questions that remain open for future exploration. From tradition to present and beyond, this work systematically reviews RAG from a perspective of the agentic era.
\end{minipage}
\end{center}


\input{sections/intro}

\input{sections/related-work}

\input{sections/indexing}

\input{sections/pre-retrieval}
\input{sections/retrieval}
\input{sections/post-retrieval}
\input{sections/orchestration}
\input{sections/agentic}
\input{sections/evaluation}
\input{sections/applications}
\input{sections/challenges}
\input{sections/conclusion}
\input{sections/ai}

\bibliographystyle{unsrtnat}
\bibliography{refs}

\end{document}

%% file: sections/intro.tex
\section{Introduction}
\label{sec:intro}

Language models (LMs) have become remarkably capable, yet adapting them to the virtually infinite world knowledge is still non-trivial. Once training ends, what a model knows is frozen into its parameters, and it can be incomplete, incorrect, or out of date. A model cannot answer questions about knowledge it was not trained on: a client's files, a product catalogue, yesterday's news. Teaching it that knowledge by fine-tuning is often not an option at all: it needs the weights, and a black-box model does not expose them. Even with the weights, the run scales with the data and degrades what the model could already do \citep{luo2023forgetting}. Nor does the model fail quietly: it invents answers \citep{ji2023hallucination}, it cannot point to the passage that supports a right answer, and it cannot make full use of its context \citep{lost-in-the-middle}. Retrieval-augmented generation (RAG) has therefore become the default way to adapt a model to knowledge that does not live in its parameters.

RAG couples a frozen language model, the generator, with external knowledge, held in corpora that can be updated and that a retriever searches: given a query, it retrieves relevant evidence and conditions generation on both the query and that evidence. Lewis et al.~\citep{lewis2020rag} introduced the term with a dense retriever and a sequence-to-sequence generator on knowledge-intensive tasks, and it has since spread to nearly every setting where an LM answers from external text.

Every step of that chain is a design decision: what is stored offline (indexing), how the query is prepared (pre-retrieval), how evidence is matched (retrieval), what evidence reaches the model and in what form (post-retrieval), and in what order the steps run (orchestration). Together they form the lifecycle this review follows. Figure~\ref{fig:lifecycle} shows the stages end to end, and Figure~\ref{fig:taxonomy} previews the taxonomy that organizes the rest of the review.

One question cuts across all five: who decides the control flow. In a pipeline the designer fixes it in advance, loops included; in an agent the model chooses at run time whether to retrieve, what to retrieve, and how often. The stages are the same in both cases, but in the second none of them is scheduled ahead of time. Because these systems differ in who decides rather than in which stage they alter, we discuss agentic RAG separately.

Existing RAG reviews fall into two groups (Section~\ref{sec:related}): those that map the pre-agent pipeline, and those that treat agentic RAG in isolation with that pipeline as background. The two are only weakly connected. A reader of the former cannot tell what RAG is like in the agentic era, what is changed by agents, and what remains unchanged; a reader of the latter has limited idea of the underlying layers beneath the agent, and what RAG is like before agents. This review connects them, and aims at a compact and coherent map walking through RAG from tradition to present and beyond. It puts both on a single lifecycle axis, so that a 2020 retriever and a 2026 agent are read at the same points in the same sequence. Table~\ref{tab:reviews} shows how this review differs from existing ones.

The review proceeds as follows. Section~\ref{sec:related} situates it against prior RAG reviews; Sections~\ref{sec:indexing}--\ref{sec:orchestration} take the five lifecycle components in turn; Section~\ref{sec:agentic} turns to agentic RAG; Section~\ref{sec:eval} covers evaluation, Section~\ref{sec:apps} applications, and Section~\ref{sec:challenges} the questions that remain open for future exploration.

\input{figures/fig_lifecycle}
\input{figures/fig_taxonomy}

\input{tables/related-work}

%% file: figures/fig_lifecycle.tex
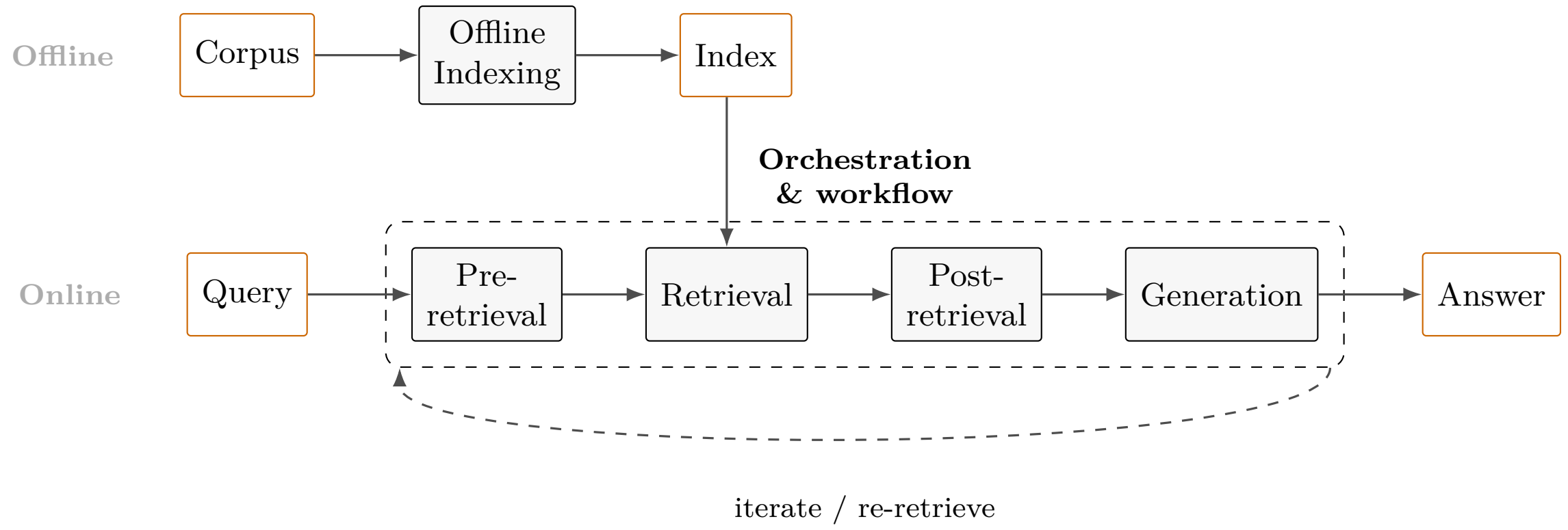
\begin{figure}[t]
\centering
\adjustbox{max width=\textwidth}{%
\begin{tikzpicture}[
  font=\footnotesize,
  stage/.style={draw, rounded corners=1pt, fill=gray!6, align=center,
                minimum height=9mm, inner sep=4pt},
  data/.style={draw=BurntOrange, fill=white, rounded corners=1pt, align=center,
               minimum height=8mm, inner sep=4pt},
  ar/.style={-{Latex[length=1.8mm]}, semithick, black!70},
  far/.style={-{Latex[length=1.8mm]}, semithick, dashed, black!70}
]
\node[data]  (corpus) {Corpus};
\node[stage, right=10mm of corpus] (idx) {Offline\\Indexing};
\node[data,  right=10mm of idx]    (index) {Index};
\node[data,  below=15mm of corpus] (query) {Query};
\node[stage, right=10mm of query]  (pre)   {Pre-\\retrieval};
\node[stage, right=8mm of pre]     (ret)   {Retrieval};
\node[stage, right=8mm of ret]     (post)  {Post-\\retrieval};
\node[stage, right=8mm of post]    (gen)   {Generation};
\node[data,  right=10mm of gen]    (ans)   {Answer};
\node[font=\scriptsize\bfseries, text=gray!65, left=5mm of corpus] {Offline};
\node[font=\scriptsize\bfseries, text=gray!65, left=5mm of query]  {Online};

\draw[ar] (corpus) -- (idx);
\draw[ar] (idx) -- (index);
\draw[ar] (query) -- (pre);
\draw[ar] (pre) -- (ret);
\draw[ar] (ret) -- (post);
\draw[ar] (post) -- (gen);
\draw[ar] (gen) -- (ans);

\draw[ar] (ret.north |- index.south) -- (ret.north);

\begin{scope}[on background layer]
  \node[draw, dashed, rounded corners=4pt, inner sep=7pt,
        fit=(pre)(gen)] (orch) {};
\end{scope}
\node[font=\scriptsize\bfseries, anchor=south, align=center] at ([yshift=1pt]orch.north)
      {Orchestration\\ \& workflow};
\draw[far] ([xshift=-4pt]orch.south east)
   .. controls ([xshift=-4pt,yshift=-9mm]orch.south east)
      and ([xshift=4pt,yshift=-9mm]orch.south west)
   .. ([xshift=4pt]orch.south west);
\node[font=\scriptsize, anchor=north] at ([yshift=-11mm]orch.south)
      {iterate / re-retrieve};
\end{tikzpicture}%
}
\caption{The RAG lifecycle. The offline row builds the index from the corpus; the online row
reuses that index for every query, passing it through pre-retrieval, retrieval, and post-retrieval
to generation. Orchestration decides whether the online stages run once, branch,
iterate, recurse, or are driven by an agent.}
\label{fig:lifecycle}
\end{figure}

%% file: figures/fig_taxonomy.tex
\providecommand{\taxref}[1]{{\scriptsize\mdseries\textcolor{black!60}{(\S\ref{#1})}}}
\begin{figure}[htbp]
\centering
\adjustbox{max width=\textwidth}{%
\hyphenpenalty=10000 \exhyphenpenalty=10000 
\begin{tikzpicture}[
  root/.style={draw=BurntOrange, rounded corners=1pt, fill=white, minimum height=8mm, inner sep=6pt, font=\bfseries\footnotesize},
  stg/.style={draw, rounded corners=1pt, fill=gray!13, minimum height=7mm, text width=3.2cm, align=center, inner sep=3pt, font=\footnotesize},
  cat/.style={draw, rounded corners=1pt, fill=gray!9, text width=3.2cm, align=center, inner sep=3pt, font=\footnotesize\bfseries},
  mlist/.style={draw, rounded corners=1pt, fill=gray!3, text width=9.2cm, align=flush left, inner sep=3pt, font=\scriptsize},
  e/.style={draw, gray!55, semithick},
  le/.style={draw, gray!40},
]
\node[mlist, at={(0,0)}, anchor=north west] (m0_0) {Late Chunking~\cite{latechunking}, LumberChunker~\cite{lumberchunker}, Meta-Chunking~\cite{metachunking}, MoC~\cite{moc}};
\node[cat, at=(m0_0.west), anchor=east, xshift=-2.5mm] (cat0) {Chunking \taxref{sec:chunking}};
\node[fit=(cat0)(m0_0), inner sep=0pt] (blk0) {};
\draw[le] (cat0.east) -- (m0_0.west);
\node[mlist, at=(m0_0.south west), anchor=north west, yshift=-4mm] (m1_0) {Doc2query~\cite{doc2query}, QBR~\cite{qbr}};
\node[cat, at=(m1_0.west), anchor=east, xshift=-2.5mm] (cat1) {Document augmentation \taxref{sec:doc-aug}};
\node[fit=(cat1)(m1_0), inner sep=0pt] (blk1) {};
\draw[le] (cat1.east) -- (m1_0.west);
\node[mlist, at=(m1_0.south west), anchor=north west, yshift=-2mm] (m2_0) {MemWalker~\cite{memwalker}, RAPTOR~\cite{raptor}, TreeRAG~\cite{treerag}, DiscoRAG~\cite{discorag}, DISRetrieval~\cite{disretrieval}, SproutRAG~\cite{sproutrag}};
\node[cat, at=(m2_0.west), anchor=east, xshift=-2.5mm] (cat2) {Trees \taxref{sec:topology}};
\node[fit=(cat2)(m2_0), inner sep=0pt] (blk2) {};
\draw[le] (cat2.east) -- (m2_0.west);
\node[mlist, at=(m2_0.south west), anchor=north west, yshift=-2mm] (m3_0) {GraphRAG~\cite{graphrag}, HippoRAG~\cite{hipporag}, LightRAG~\cite{lightrag}, AutoGraph-R1~\cite{autographr1}, SubgraphRAG~\cite{subgraphrag}, MiniRAG~\cite{minirage}, SURGE~\cite{surge}, T-Retriever~\cite{tretriever}, LinearRAG~\cite{LinearRAG}, TIGRAG~\cite{TIGRAG}};
\node[cat, at=(m3_0.west), anchor=east, xshift=-2.5mm] (cat3) {Graphs \taxref{sec:topology}};
\node[fit=(cat3)(m3_0), inner sep=0pt] (blk3) {};
\draw[le] (cat3.east) -- (m3_0.west);
\node[mlist, at=(m3_0.south west), anchor=north west, yshift=-2mm] (m3_h) {HyperGraphRAG~\cite{hypergraphrag}, PropRAG~\cite{proprag}};
\node[cat, at=(m3_h.west), anchor=east, xshift=-2.5mm] (cat3h) {Hypergraphs \taxref{sec:topology}};
\node[fit=(cat3h)(m3_h), inner sep=0pt] (blk3h) {};
\draw[le] (cat3h.east) -- (m3_h.west);
\node[mlist, at=(m3_h.south west), anchor=north west, yshift=-2mm] (m3_o) {OG-RAG~\cite{ograg}, MultiCube-RAG~\cite{multicuberag}};
\node[cat, at=(m3_o.west), anchor=east, xshift=-2.5mm] (cat3o) {Ontology \taxref{sec:topology}};
\node[fit=(cat3o)(m3_o), inner sep=0pt] (blk3o) {};
\draw[le] (cat3o.east) -- (m3_o.west);
\node[mlist, at=(m3_o.south west), anchor=north west, yshift=-2mm] (m4_0) {SKR~\cite{skr}, Rowen~\cite{rowen}, Adaptive-RAG~\cite{adaptiverag}, RAGRouter~\cite{ragrouter}, UAR~\cite{uar}, Flare-Aug~\cite{flareaug}};
\node[cat, at=(m4_0.west), anchor=east, xshift=-2.5mm] (cat4) {Routing \taxref{sec:routing}};
\node[fit=(cat4)(m4_0), inner sep=0pt] (blk4) {};
\draw[le] (cat4.east) -- (m4_0.west);
\node[mlist, at=(m4_0.south west), anchor=north west, yshift=-2mm] (m5_0) {DSP~\cite{dsp}, QE~\cite{qe}, HyDE~\cite{hyde}, Query2Doc~\cite{query2doc}, Query Rewriting~\cite{qrewrite}, GAR-meets-RAG~\cite{rrr}, RQ-RAG~\cite{rqrag}, UniRAG~\cite{unirag}};
\node[cat, at=(m5_0.west), anchor=east, xshift=-2.5mm] (cat5) {Query augmentation \taxref{sec:query-aug}};
\node[fit=(cat5)(m5_0), inner sep=0pt] (blk5) {};
\draw[le] (cat5.east) -- (m5_0.west);
\node[mlist, at=(m5_0.south west), anchor=north west, yshift=-2mm] (m6_0) {BM25~\cite{bm25}};
\node[cat, at=(m6_0.west), anchor=east, xshift=-2.5mm] (cat6) {Sparse \taxref{sec:ret-sparse}};
\node[fit=(cat6)(m6_0), inner sep=0pt] (blk6) {};
\draw[le] (cat6.east) -- (m6_0.west);
\node[mlist, at=(m6_0.south west), anchor=north west, yshift=-2mm] (m7_0) {DPR~\cite{dpr}, ColBERT~\cite{colbert}, AAR~\cite{aar}, Multi-head RAG~\cite{mrag}};
\node[cat, at=(m7_0.west), anchor=east, xshift=-2.5mm] (cat7) {Dense \taxref{sec:ret-dense}};
\node[fit=(cat7)(m7_0), inner sep=0pt] (blk7) {};
\draw[le] (cat7.east) -- (m7_0.west);
\node[mlist, at=(m7_0.south west), anchor=north west, yshift=-2mm] (m8_0) {sparse--dense fusion~\cite{luan2021sparse, wang2024searching}, RRF~\cite{rrf}};
\node[cat, at=(m8_0.west), anchor=east, xshift=-2.5mm] (cat8) {Hybrid \taxref{sec:ret-hybrid}};
\node[fit=(cat8)(m8_0), inner sep=0pt] (blk8) {};
\draw[le] (cat8.east) -- (m8_0.west);
\node[mlist, at=(m8_0.south west), anchor=north west, yshift=-2mm] (m9_0) {MemWalker~\cite{memwalker}, RAPTOR~\cite{raptor}, TreeRAG~\cite{treerag}, DISRetrieval~\cite{disretrieval}, T-Retriever~\cite{tretriever}, HippoRAG~\cite{hipporag}, GraphRAG~\cite{graphrag}, LightRAG~\cite{lightrag}, SubgraphRAG~\cite{subgraphrag}, PropRAG~\cite{proprag}, HyperGraphRAG~\cite{hypergraphrag}, TaSR-RAG~\cite{tasrrag}, PyRAG~\cite{pyrag}};
\node[cat, at=(m9_0.west), anchor=east, xshift=-2.5mm] (cat9) {Structured \taxref{sec:ret-structured}};
\node[fit=(cat9)(m9_0), inner sep=0pt] (blk9) {};
\draw[le] (cat9.east) -- (m9_0.west);
\node[mlist, at=(m9_0.south west), anchor=north west, yshift=-2mm] (m10_0) {Re2G~\cite{re2g}, GenRT~\cite{genrt}, RankGPT~\cite{rankgpt}, PRP~\cite{prp}, ChunkRAG~\cite{chunkrag}, RankRAG~\cite{rankrag}, SparseRAG~\cite{sparserag}, DynamicRAG~\cite{dynamicrag}, EBCAR~\cite{ebcar}, CoRank~\cite{corank}, SANE~\cite{sane}};
\node[cat, at=(m10_0.west), anchor=east, xshift=-2.5mm] (cat10) {Reranking \taxref{sec:reranking}};
\node[fit=(cat10)(m10_0), inner sep=0pt] (blk10) {};
\draw[le] (cat10.east) -- (m10_0.west);
\node[mlist, at=(m10_0.south west), anchor=north west, yshift=-2mm] (m11_0) {FilCo~\cite{filco}, PRCA~\cite{prca}, Token Elimination~\cite{tokenelemination}, RECOMP~\cite{recomp}, Provence~\cite{provence}, Corrective RAG~\cite{crag}, StructRAG~\cite{structrag}, DTCRS~\cite{dtcrs}, LongRefiner~\cite{longrefiner}, SARG~\cite{sarg}, RAS~\cite{ras}, SLIDERS~\cite{sliders}, RAG-CSM~\cite{RAG-CSM}};
\node[cat, at=(m11_0.west), anchor=east, xshift=-2.5mm] (cat11) {Refinement \taxref{sec:refinement}};
\node[fit=(cat11)(m11_0), inner sep=0pt] (blk11) {};
\draw[le] (cat11.east) -- (m11_0.west);
\node[mlist, at=(m11_0.south west), anchor=north west, yshift=-2mm] (m12_0) {In-Context RALM~\cite{incontextralm}, RA-DIT~\cite{radit}, FiD~\cite{fid}, FiD-Light~\cite{fidlight}, COCOM~\cite{cocom}, xRAG~\cite{xrag}, REFRAG~\cite{refrag}, Retro~\cite{retro}, kNN-LM~\cite{knnlm}, Efficient kNN-LM~\cite{efficientknnlm}, REST~\cite{rest}, RePlug~\cite{replug}, P-RAG~\cite{prag}};
\node[cat, at=(m12_0.west), anchor=east, xshift=-2.5mm] (cat12) {Fusion \taxref{sec:fusion}};
\node[fit=(cat12)(m12_0), inner sep=0pt] (blk12) {};
\draw[le] (cat12.east) -- (m12_0.west);
\node[mlist, at=(m12_0.south west), anchor=north west, yshift=-2mm] (m13_0) {RAG~\cite{lewis2020rag}, Re2G~\cite{re2g}, MedRGAG~\cite{medrgag}};
\node[cat, at=(m13_0.west), anchor=east, xshift=-2.5mm] (cat13) {Linear \taxref{sec:orch-linear}};
\node[fit=(cat13)(m13_0), inner sep=0pt] (blk13) {};
\draw[le] (cat13.east) -- (m13_0.west);
\node[mlist, at=(m13_0.south west), anchor=north west, yshift=-2mm] (m14_0) {GRG~\cite{grg}, BlendFilter~\cite{blendfilter}, RePlug~\cite{replug}};
\node[cat, at=(m14_0.west), anchor=east, xshift=-2.5mm] (cat14) {Branching \taxref{sec:orch-branch}};
\node[fit=(cat14)(m14_0), inner sep=0pt] (blk14) {};
\draw[le] (cat14.east) -- (m14_0.west);
\node[mlist, at=(m14_0.south west), anchor=north west, yshift=-2mm] (m15_0) {IRCoT~\cite{ircot}, Iter-RetGen~\cite{iterretgen}, FLARE~\cite{flare}, Self-Memory~\cite{selfmem}, ReflectiveRAG~\cite{reflectiverag}, RAS~\cite{ras}, Q-RAG~\cite{Q-RAG}};
\node[cat, at=(m15_0.west), anchor=east, xshift=-2.5mm] (cat15) {Iterative \taxref{sec:orch-iter}};
\node[fit=(cat15)(m15_0), inner sep=0pt] (blk15) {};
\draw[le] (cat15.east) -- (m15_0.west);
\node[mlist, at=(m15_0.south west), anchor=north west, yshift=-2mm] (m16_0) {ToC~\cite{toc}, RA-ISF~\cite{raisf}, TreeRare~\cite{treerare}};
\node[cat, at=(m16_0.west), anchor=east, xshift=-2.5mm] (cat16) {Recursive \taxref{sec:orch-rec}};
\node[fit=(cat16)(m16_0), inner sep=0pt] (blk16) {};
\draw[le] (cat16.east) -- (m16_0.west);
\node[mlist, at=(m16_0.south west), anchor=north west, yshift=-2mm] (m17_0) {Self-RAG~\cite{selfrag}, MA-RAG~\cite{marag}, Graph-R1~\cite{graphr1}, Search-R1~\cite{searchr1}, Search-o1~\cite{searcho1}, DeepRAG~\cite{deeprag}, HiPRAG~\cite{hiprag}, Search-R2~\cite{searchr2}, JADE~\cite{jade}, A-RAG~\cite{arag}, DocNavRAG~\cite{docnavrag}, VecTree-RAG~\cite{vectreerag}, CoRAG~\cite{CoRAG}, LogicRAG~\cite{LogicRAG}, MMOA-RAG~\cite{MMOA-RAG}, TeaRAG~\cite{TeaRAG}, IGPO~\cite{igpo}};
\node[cat, at=(m17_0.west), anchor=east, xshift=-2.5mm] (cat17) {Agentic \taxref{sec:agentic}};
\node[fit=(cat17)(m17_0), inner sep=0pt] (blk17) {};
\draw[le] (cat17.east) -- (m17_0.west);
\node[fit=(blk0)(blk1)(blk2)(blk3)(blk3h)(blk3o), inner sep=0pt] (fs0) {};
\node[stg, left=6mm of fs0] (s0) {Offline Indexing \taxref{sec:indexing}};
\node[fit=(blk4)(blk5), inner sep=0pt] (fs1) {};
\node[stg, left=6mm of fs1] (s1) {Pre-retrieval \taxref{sec:pre-retrieval}};
\node[fit=(blk6)(blk7)(blk8)(blk9), inner sep=0pt] (fs2) {};
\node[stg, left=6mm of fs2] (s2) {Retrieval \taxref{sec:retrieval}};
\node[fit=(blk10)(blk11)(blk12), inner sep=0pt] (fs3) {};
\node[stg, left=6mm of fs3] (s3) {Post-retrieval \taxref{sec:post-retrieval}};
\node[fit=(blk13)(blk14)(blk15)(blk16)(blk17), inner sep=0pt] (fs4) {};
\node[stg, left=6mm of fs4] (s4) {Orchestration \taxref{sec:orchestration}};
\node[fit=(s0)(s1)(s2)(s3)(s4), inner sep=0pt] (fall) {};
\node[root, left=7mm of fall] (root) {RAG};
\draw[e] (root.east) -- (s0.west);
\draw[le] (s0.east) -- (blk0.west);
\draw[le] (s0.east) -- (blk1.west);
\draw[le] (s0.east) -- (blk2.west);
\draw[le] (s0.east) -- (blk3.west);
\draw[le] (s0.east) -- (blk3h.west);
\draw[le] (s0.east) -- (blk3o.west);
\draw[e] (root.east) -- (s1.west);
\draw[le] (s1.east) -- (blk4.west);
\draw[le] (s1.east) -- (blk5.west);
\draw[e] (root.east) -- (s2.west);
\draw[le] (s2.east) -- (blk6.west);
\draw[le] (s2.east) -- (blk7.west);
\draw[le] (s2.east) -- (blk8.west);
\draw[le] (s2.east) -- (blk9.west);
\draw[e] (root.east) -- (s3.west);
\draw[le] (s3.east) -- (blk10.west);
\draw[le] (s3.east) -- (blk11.west);
\draw[le] (s3.east) -- (blk12.west);
\draw[e] (root.east) -- (s4.west);
\draw[le] (s4.east) -- (blk13.west);
\draw[le] (s4.east) -- (blk14.west);
\draw[le] (s4.east) -- (blk15.west);
\draw[le] (s4.east) -- (blk16.west);
\draw[le] (s4.east) -- (blk17.west);
\end{tikzpicture}%
}
\caption{The five-component taxonomy of the RAG lifecycle. Section numbers (\S) point to the section or subsection that discusses each stage and category.}
\label{fig:taxonomy}
\end{figure}
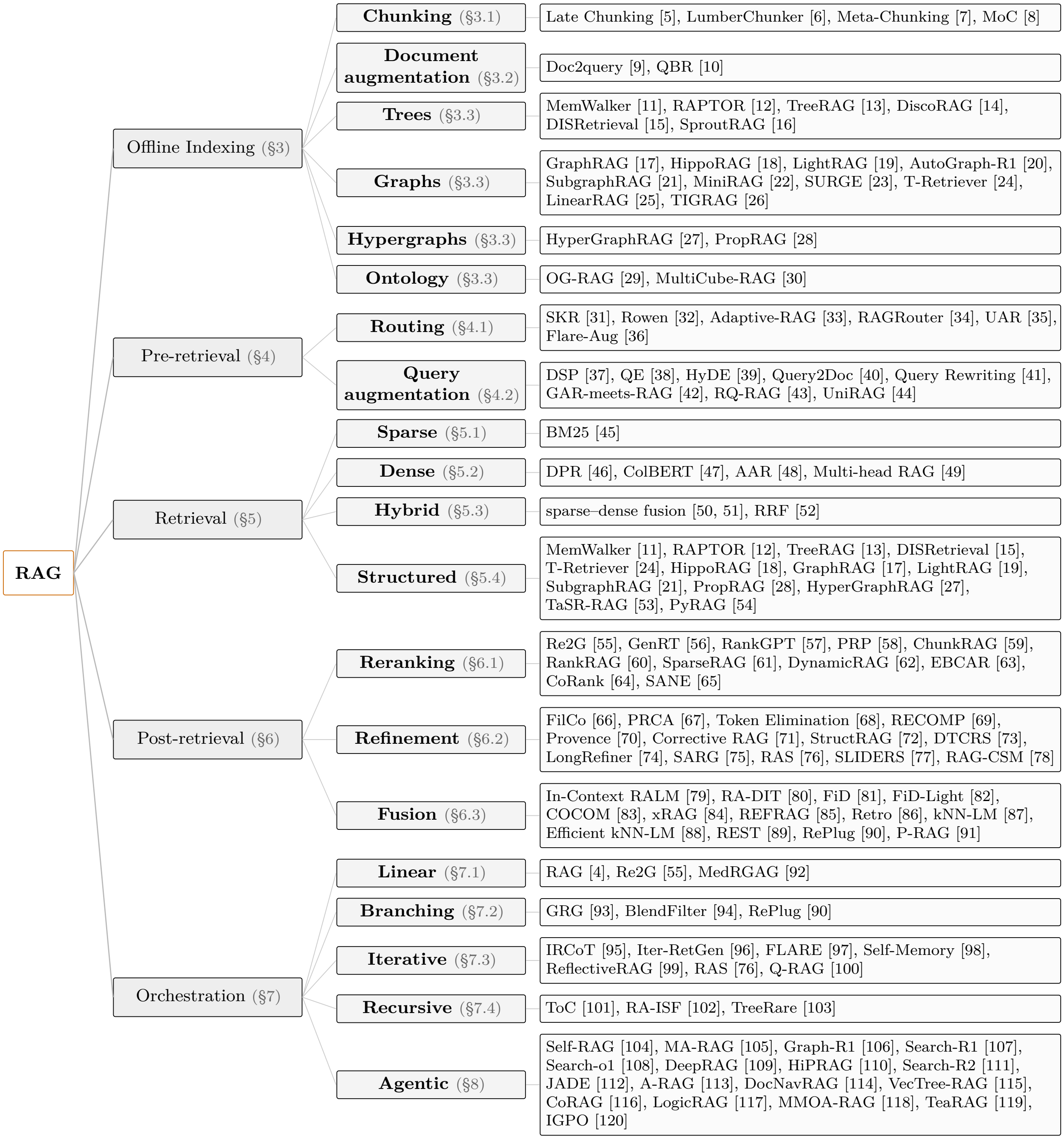

%% file: tables/related-work.tex
\begin{table}[t]
\centering
\footnotesize
\setlength{\tabcolsep}{4pt}
\renewcommand{\arraystretch}{1.08}
\begin{tabularx}{\textwidth}{@{}l l >{\raggedright\arraybackslash}X c c@{}}
\toprule
\textbf{Review} & \textbf{Scope} & \textbf{Organizing axis} & \textbf{Traditional RAG} & \textbf{Agentic RAG} \\
\midrule
\citet{li2022ragtextgen}            & General        & Text generation            & \cmark & \xmark \\
\citet{fan2024ragmeeting}           & General        & Pipeline                   & \cmark & \xmark \\
\citet{gao2023ragreview}            & General        & Paradigms                  & \cmark & \xmark \\
\citet{rag2025comprehensive}        & General        & Architecture focuses       & \cmark & \xmark \\
\citet{cheng2025korag}              & General        & Techniques                 & \cmark & \xmark \\
\citet{wu2026nlp}                   & General        & Pipeline                   & \cmark & \xmark \\
\citet{huang2026ragtext}            & General        & Pipeline                   & \cmark & \xmark \\
\citet{peng2024graphrag}            & Graph RAG      & Pipeline                   & \pmark & \xmark \\
\citet{han2025graphrag}             & Graph RAG      & Knowledge operations       & \pmark & \xmark \\
\citet{singh2026agentic}            & Agentic RAG    & Agent techniques           & \xmark & \cmark \\
\citet{mishra2026sok}               & Agentic RAG    & Architectures and patterns & \xmark & \cmark \\
\citet{yu2024eval}                  & Evaluation     & Evaluation components      & \cmark & \xmark \\
\citet{ahmed2026trustworthy}        & Trustworthy    & Trustworthy dimensions     & \cmark & \pmark \\
\citet{mei2025multimodal}           & Multi-modality & Pipeline                   & \cmark & \pmark \\
\citet{zhao2026aigc}                & Multi-modality & Modality                   & \cmark & \xmark \\
\midrule
\textbf{Ours} & \textbf{General} & \textbf{RAG lifecycle} & \cmark & \cmark \\
\bottomrule
\end{tabularx}
\caption{Representative RAG reviews and how this review is positioned.
\cmark{} = covered; \pmark{} = partially covered; \xmark{} = not addressed.}
\label{tab:reviews}
\end{table}

%% file: sections/related-work.tex
\section{Related Work}
\label{sec:related}

RAG has attracted a large and rapidly growing review literature spanning multiple topics. In this section, we categorize and introduce the most representative and impactful RAG reviews.

\paragraph{General RAG reviews.}
These reviews focus on the general RAG literature. To the best of our knowledge, \citet{li2022ragtextgen} is the earliest review on RAG and consolidate pre-LLM retrieval-augmented text generation, focusing on sources, metrics, models, and tasks. With the rise of large-scale LMs, general reviews pivot to the RAG--LM interface. \citet{fan2024ragmeeting} review retrieval-augmented language models with a special focus on techniques of retrieval, generation, and training. \citet{gao2023ragreview} introduce three paradigms with respect to the development of RAG approaches: naive RAG, advanced RAG, and modular RAG. \citet{rag2025comprehensive} organize RAG architectures based on their focuses: retriever-centric, generator-centric, hybrid, and robustness-oriented. \citet{cheng2025korag} bring a linearized review of both basic RAG techniques and advanced RAG approaches. \citet{wu2026nlp} review RAG from a pipeline perspective with an emphasis on retrieval fusion (query-based, logits-based, latent, and parametric). \citet{huang2026ragtext} adopt a four-stage pipeline (pre-retrieval,
retrieval, post-retrieval, generation) to organize RAG approaches. Our organization, focusing on the whole lifecycle of RAG, is inspired by \citet{huang2026ragtext}, but we treat indexing and orchestration as first-class components and additionally discuss the recent stream of agentic RAG that has appeared since those reviews.

\paragraph{Graph RAG reviews.}
A fast-growing family studies explicit graph structure, which is widely adopted. \citet{peng2024graphrag} give
a comprehensive overview for Graph RAG, focusing on indexing, retrieval, generation, and training. \citet{han2025graphrag} introduce techniques for organization, retrieval, and integration of knowledge in Graph RAG. In this review, we treat graph methods as one option inside indexing
(Section~\ref{sec:indexing}) rather than as a separate first-class paradigm.

\paragraph{Agentic RAG reviews.}
A dedicated agentic RAG literature emerged with the development of agents. \citet{singh2026agentic} review agentic RAG with an emphasis on techniques in the agent literature (e.g., planning, memory, reflection), while \citet{mishra2026sok} systematize architectural components and workflow design patterns. These reviews are deep on autonomy but largely assume the traditional RAG pipeline as background. Meanwhile, there have been a lot of novel and modern agentic approaches for RAG emerging after their publications, which have never been reviewed. Although agentic RAG can be categorized as a type of orchestration (Section~\ref{sec:orchestration}), we discuss it in a separate section (Section~\ref{sec:agentic}) since its paradigm is significantly different from prior ones.

\paragraph{Reviews of specific facets.}
There are a few reviews focusing on different perspectives of RAG like evaluation, trustworthiness, and multi-modality. \citet{yu2024eval} review evaluation for RAG, focusing on target, dataset, and metric. \citet{ahmed2026trustworthy} review trustworthy RAG across multiple dimensions of trustworthiness. \citet{mei2025multimodal} review multimodal RAG, while \citet{zhao2026aigc} review RAG for AI-generated content across modalities. We primarily review the general aspects of RAG: methods, evaluation, and application.

\paragraph{How this review differs.}
Existing reviews typically either (i) focus on traditional pre-agent RAG, or (ii) dive deep
into agentic RAG in isolation, so the two worlds are only weakly connected in prior review
treatments. Our goal is a review of RAG \emph{in the agentic era}: we cover both the
traditional pipeline and modern agentic workflows, organized around a single end-to-end
lifecycle axis.\footnote{We define RAG lifecycle as the pipeline containing indexing,
pre-retrieval, retrieval, and post-retrieval (which includes generation) stages, and an orchestration component controlling over these stages.} Indexing and
orchestration are treated as first-class components, and the agentic literature published
since those reviews is included. The account
is compact but systematic, spanning the whole lifecycle and placing classic pipeline design
under agent-era constraints; Table~\ref{tab:reviews} compares the closest reviews along these
lines. What we review is textual RAG. Multimodal retrieval (tables, databases, images,
audio, video) and production system design fall outside it.

%% file: sections/indexing.tex
\section{Offline Indexing}
\label{sec:indexing}

Indexing is an offline stage that processes the corpus before any queries: how documents are split, enriched, and organized so that a retriever can later find the right evidence. We cover three families of indexing techniques: chunking, document augmentation, and advanced structures (Figure~\ref{fig:indexing}).

\input{figures/fig_indexing}

\subsection{Chunking}
\label{sec:chunking}

A corpus is typically a set of long documents, but retrieval needs units of a workable size, which necessitates chunking. The naive solution, a fixed window of $n$ tokens with some
overlap, is fast and predictable but severs content mid-thought, and a
chunk's embedding then reflects a fragment rather than a coherent idea. Late chunking makes the embedding contextualized, and adaptive chunking makes chunks coherent.

\paragraph{Late chunking.}
Conventional pipelines chunk first and embed each chunk in isolation, so a chunk embedding
never sees the surrounding document. Late chunking \citep{latechunking} inverts the order:
embed the entire long document with a long-context encoder, then pool the token embeddings
into chunk vectors. Each chunk embedding is thus contextualized by its document, at the cost
of one long-context forward pass per document.

\paragraph{Adaptive chunking.}
Rather than fixing a length, several methods decide boundaries adaptively.
LumberChunker \citep{lumberchunker} feeds successive windows to an LM and asks where the
next semantic split falls, producing variable-length, content-defined segments.
Meta-Chunking \citep{metachunking} uses a cheaper signal: it cuts where model uncertainty
or a perplexity threshold indicates a semantic boundary, avoiding an LM call per decision.
MoC \citep{moc} first classifies the appropriate chunking granularity and then routes to a matching splitting rule. The cost of adaptivity is typically an extra model pass of the corpus.

\subsection{Document Augmentation}
\label{sec:doc-aug}

Raw documents can be imperfect retrieval targets: a document states facts, while a query asks
about them, and the two are distributionally misaligned. Document augmentation
narrows this gap by enriching documents with query-like text. Doc2query \citep{doc2query} generates hypothetical queries for each document and appends
them before indexing, so a query is more likely to match. A later
enterprise-oriented method \citep{qbr} splits a document into atomic units, generates several
questions per unit, and retrieves against the questions while mapping hits back to the unit.
This is more faithful to multi-interest documents, though the question
generation can still lose information.

\subsection{Advanced Structures}
\label{sec:topology}

Text is unstructured, but the knowledge inside it is not: sections contain subsections,
arguments have premises and conclusions, and entities relate to one another. A large family
of methods imposes an explicit \emph{structure} --- a tree, a graph, or a hypergraph --- so that retrieval can follow hierarchies and relationships rather than rely on flat similarity.

\paragraph{Trees.}
Trees express a hierarchical structure. MemWalker
\citep{memwalker} builds a summary tree and lets an LM walk it interactively, expanding
subtrees on demand. RAPTOR \citep{raptor} is the canonical method: it clusters chunks,
summarizes each cluster, then recursively clusters and summarizes the summaries, forming a
tree retrieved top-down. 
SproutRAG \citep{sproutrag} adds attention-guided
tree search with progressive embeddings. TreeRAG \citep{treerag} uses the document's own section hierarchy.
Discourse-aware methods build a tree from rhetorical structure: DiscoRAG
\citep{discorag} and DISRetrieval \citep{disretrieval} use discourse relations as retrieval paths.

\paragraph{Graphs.}
Graphs capture relationships that trees cannot, and Graph RAG has been a widely-studied paradigm \cite{han2025graphrag}. Earlier graph-augmented language
models such as SURGE \citep{surge} encoded a knowledge graph with a graph network.
Microsoft's GraphRAG \citep{graphrag}
extracts entities and relations into a knowledge graph, detects communities, summarizes
them, and answers from both community summaries and entity paths. HippoRAG
\citep{hipporag} is inspired by the hippocampus: it extracts triples, links the query to
graph nodes, and retrieves by Personalized PageRank, so activating one memory activates its
neighbours. 
LightRAG \citep{lightrag} features simpler construction and leverages
both low-level and high-level keys to boost retrieval. AutoGraph-R1
\citep{autographr1} trains graph construction with reinforcement learning from downstream accuracy. A hybrid tree-on-graph method,
T-Retriever \citep{tretriever}, builds a hierarchy over a textual graph. Several works reduce cost or pursue simplicity: SubgraphRAG \citep{subgraphrag} argues that simple subgraph retrieval with an LM is competitive, MiniRAG \citep{minirage} targets extreme simplicity,  LinearRAG \citep{LinearRAG} explores a relation-free hierarchical graph linking passages and sentences with co-occurring entities, and TIGRAG \citep{TIGRAG} performs retrieval based on simple token co-occurrence graphs that are cheap to construct.  

\paragraph{Hypergraphs.}
Ordinary edges are binary, but many facts are $n$-ary. HyperGraphRAG
\citep{hypergraphrag} represents such facts as hyperedges (e.g., a clinical definition that
binds three entities at once), stored as nodes for compatibility with graph databases.
PropRAG \citep{proprag} builds a hypergraph over extracted propositions and retrieves by beam
search over proposition paths.

\paragraph{Ontology-guided Structures.}
Ontology-guided RAG, being increasingly popular in enterprise application domains recently due to its reliability and faithfulness, pushes further. OG-RAG
\citep{ograg} fills documents into a domain ontology so that types and fields become typed
edges. MultiCube-RAG \citep{multicuberag} maps entities into an ontology cube of typed
dimensions. The appeal is typed, rule-constrained construction rather than aimless
extraction; the cost is that ontologies must be authored by hand and maintained as the domain evolves.

\subsection{Discussion}
\label{sec:indexing-disc}

Chunking, augmentation, and structuring all spend offline time to change what the online budget
buys, though the first two are cheap next to the third. An index is a bet that construction time
returns either higher quality at the same online cost, or the same quality at a lower cost. Indexing also decides something before a query exists, and cannot learn at index time
the correctness of the decision: chunking fixes where evidence may sit, augmentation fixes the
wording that reaches it, structure fixes the paths retrieval may follow. Only the first
commitment is unavoidable, so the other two carry the burden of proof: augmentation pays when
the corpus is stable and the form of its queries is predictable, and structure pays only when
the task's reasoning follows the structure, which must then supply information that similarity
cannot, a bar that rises with every new embedding model. Ontology-guided construction, now common in practice, pushes this
further: documents are fitted to a domain ontology, so an edge carries a type rather than
whatever an extractor guessed. That is also its cost, the hand-authored part: an ontology has to
be written and then rewritten as the domain changes. The question worth asking is therefore how
much of this guessing can be moved to retrieval time. On the evidence of this section, very
little: every method here decides at index time, and the corpus is fixed before the first query
arrives.

%% file: figures/fig_indexing.tex
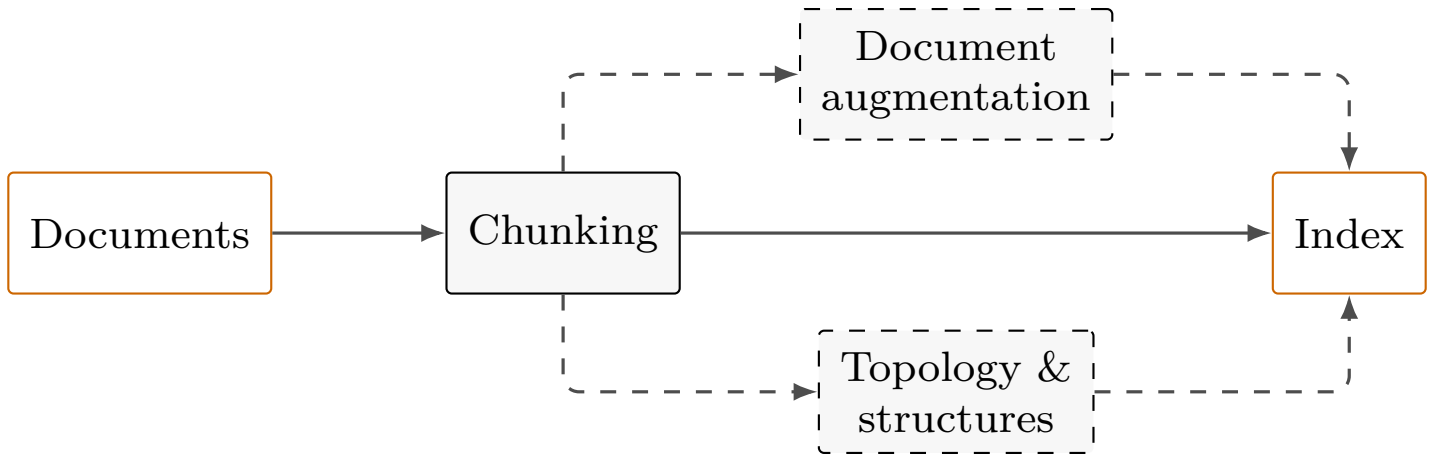
\begin{figure}[t]
\centering
\adjustbox{max width=\textwidth}{%
\begin{tikzpicture}[font=\scriptsize,
  d/.style={draw=BurntOrange, rounded corners=1pt, fill=white, minimum height=8mm, inner sep=4pt, align=center},
  s/.style={draw, rounded corners=1pt, fill=gray!6, minimum height=8mm, inner sep=4pt, align=center},
  o/.style={draw, dashed, rounded corners=1pt, fill=gray!6, minimum height=8mm, inner sep=4pt, align=center},
  a/.style={-{Latex[length=1.6mm]}, semithick, black!70},
  oa/.style={-{Latex[length=1.6mm]}, semithick, dashed, black!70},
  el/.style={rounded corners=3pt}
]
\node[d] (doc)   at (0,0)   {Documents};
\node[s] (chunk) at (2.8,0) {Chunking};
\node[d] (idx)   at (8.0,0) {Index};
\node[o] (aug)   at (5.4,1.05)  {Document\\augmentation};
\node[o] (topo)  at (5.4,-1.05) {Topology \&\\structures};
\draw[a] (doc)--(chunk);
\draw[a] (chunk)--(idx);
\draw[oa,el] (chunk.north) |- (aug.west);
\draw[oa,el] (aug.east) -| (idx.north);
\draw[oa,el] (chunk.south) |- (topo.west);
\draw[oa,el] (topo.east) -| (idx.south);
\end{tikzpicture}%
}
\caption{Offline indexing. Documents are chunked and an index is built (solid); the two
middle steps are optional and corpus-specific (dashed). Augmentation enriches what is
indexed, while trees, graphs and hypergraphs change how it is organized.}
\label{fig:indexing}
\end{figure}

%% file: sections/pre-retrieval.tex
\section{Pre-retrieval Processing}
\label{sec:pre-retrieval}

After receiving the query, there can be a few pre-retrieval processes to decide what the retriever is asked to do. It has two families: routing,
which picks a retrieval strategy per query, and query augmentation, which reshapes the
query itself (Figure~\ref{fig:preretrieval}).

\input{figures/fig_preretrieval}

\subsection{Routing}
\label{sec:routing}

Queries can be very diverse. Some need no retrieval at all, some need higher budget, while some need different models. Routing chooses a retrieval strategy per query: whether to retrieve at all, how large a
lookup budget to spend, and which model to run.

\paragraph{Whether to retrieve.}
Self-knowledge routing \citep{skr} asks the model whether it already knows the answer and
retrieves only if it does not. The approach is cheap, but it inherits a familiar failure mode:
models are poor judges of their own knowledge boundary. Rowen \citep{rowen} targets exactly this gap with consistency, a more
robust signal: generate an initial answer, perturb the query
(paraphrase, other languages, other models), and measure agreement; high agreement suggests
the model knows, low agreement suggests retrieval is needed. Unified Active Retrieval \citep{uar} deploys a few classifiers to decide whether to retrieve from multiple perspectives.

\paragraph{How much to retrieve.}
Adaptive-RAG \citep{adaptiverag} trains a classifier that sorts queries into three regimes:
trivial (answer directly), simple (one retrieval), and complex multi-hop (multi-round
retrieval). Flare-Aug \citep{flareaug} exposes this trade-off to the user as an
accuracy-versus-cost dial.

\paragraph{Which model.}
RAGRouter \citep{ragrouter} routes a query to one of several retrieval-augmented language
models, learned from models' downstream performance. This can be useful when a system mixes
models of different cost and competence.

\subsection{Query Augmentation}
\label{sec:query-aug}

User queries are short, ambiguous, and often not phrased like the evidence that answers
them. Query augmentation rewrites, expands, or decomposes the query before retrieval.

\paragraph{Expansion.}
Expansion adds terms to improve recall. Prompting an LM for a pseudo answer
\citep{qe} is the simplest form. HyDE \citep{hyde} generates a hypothetical answer and
retrieves with it, on the intuition that a document looks more like an answer than like a
question. Query2Doc \citep{query2doc} is the mirror image of Doc2Query: it generates a
pseudo-document from the query and retrieves with both. The two together nicely illustrate
the alignment problem: queries and documents can be brought closer from either side.

\paragraph{Rewriting.}
Queries can be imperfect for direct retrieval, and rewriting restates it. Query Rewriting \citep{qrewrite} trains such a rewriter in front
of a frozen retriever and reader, and finds
that a complex question is often a poor search query, so rewriting pays off most where
queries and evidence are least alike. GAR-meets-RAG \citep{rrr} instead closes a loop,
alternating rewriting and retrieval over several rounds before reranking, so that recall and precision are bought in separate stages.
UniRAG \citep{unirag} unifies various operations (no augmentation, paraphrasing,
expansion, abstraction) under one query-understanding module, emitting a
special token that selects the operation.

\paragraph{Decomposition.}
Complex queries may need subquery decomposition to be properly answered.
DSP \citep{dsp} prompts models to decompose queries with in-context demonstrations.
RQ-RAG \citep{rqrag} learns to refine queries through decomposition for multi-hop questions.

\subsection{Discussion}
\label{sec:pre-retrieval-disc}

Routing and augmentation both work the query side of the match, and both decide before any evidence is seen: routing picks the machinery, augmentation rewrites the text that enters it. Neither can observe its own error: a decision that spends too little and one that rewrites too freely both return an answer, and the loss surfaces as a weaker answer with no trace of where it was lost. Both also decide from an estimate of what the model knows rather than from a check of it, and the estimate is what the decision inherits. The two families then differ in which way they bet: routing bets that spending less costs nothing, augmentation bets that spending more pays. Augmentation's risk is also closer to its gain than it looks, since the text it adds is generated and the retriever will match it whether or not the corpus supports it. What is missing is a check that does not cost a second run, since knowing whether a decision was wrong means running the path it did not take.

%% file: figures/fig_preretrieval.tex
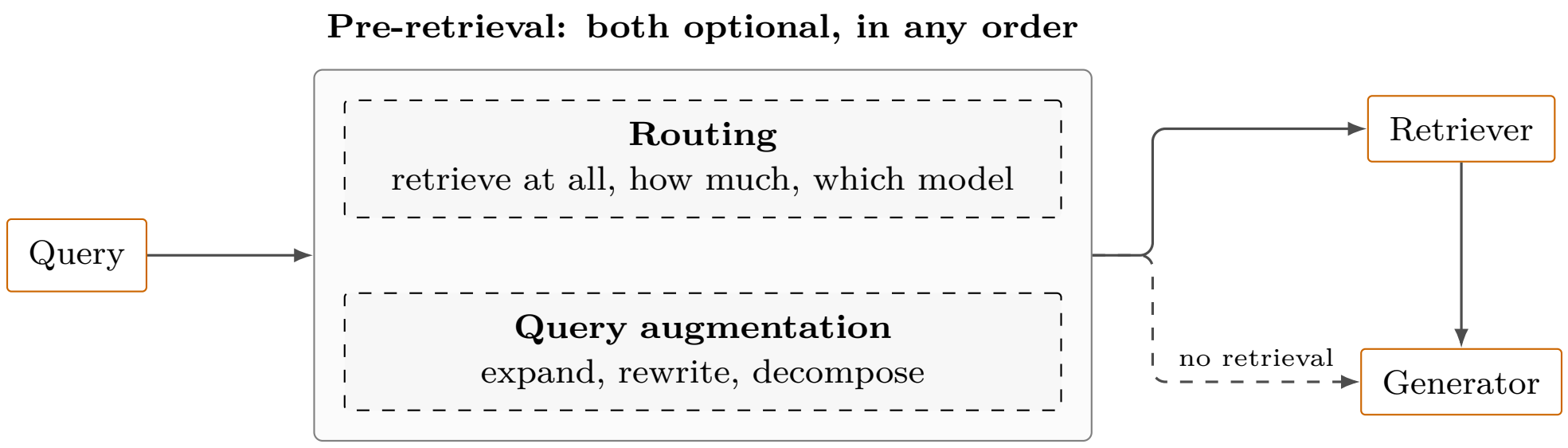
\begin{figure}[t]
\centering
\adjustbox{max width=\textwidth}{%
\begin{tikzpicture}[font=\scriptsize,
  d/.style={draw=BurntOrange, rounded corners=1pt, fill=white, inner sep=5pt, align=center},
  o/.style={draw, dashed, rounded corners=1pt, fill=gray!6, inner sep=5pt, align=center, text width=5.6cm},
  a/.style={-{Latex[length=1.6mm]}, semithick, black!70},
  oa/.style={-{Latex[length=1.6mm]}, semithick, dashed, black!70},
  el/.style={rounded corners=3pt}
]
\node[d] (q) at (0,0) {Query};
\node[o] (route) at (5.2,0.80) {{\bfseries Routing}\\[1pt]retrieve at all, how much, which model};
\node[o] (aug)   at (5.2,-0.80) {{\bfseries Query augmentation}\\[1pt]expand, rewrite, decompose};
\begin{scope}[on background layer]
  \node[draw=black!50, rounded corners=2pt, fill=gray!3, inner sep=7pt, fit=(route)(aug)] (pre) {};
\end{scope}
\node[font=\scriptsize\bfseries, anchor=south] at ([yshift=2pt]pre.north)
      {Pre-retrieval: both optional, in any order};
\node[d] (ret) at (11.5,1.05) {Retriever};
\node[d] (gen) at (11.5,-1.05) {Generator};
\draw[a] (q.east) -- (pre.west);
\coordinate (f) at ([xshift=5mm]pre.east);
\draw[a,el]  (pre.east) -- (f) |- (ret.west);
\draw[oa,el] (pre.east) -- (f) |- (gen.west);
\node[font=\tiny, above] at ($(f |- gen.west)!0.5!(gen.west)$) {no retrieval};
\draw[a] (ret.south) -- (gen.north);
\end{tikzpicture}%
}
\caption{Pre-retrieval. Two families of preparation, both optional and applicable in either
order (dashed). Routing picks a retrieval strategy per query and may answer without
retrieving; query augmentation reshapes the query before it is retrieved.}
\label{fig:preretrieval}
\end{figure}

%% file: sections/retrieval.tex
\section{Retrieval}
\label{sec:retrieval}

The retrieval stage goes to the corpus: it searches the units built in Section~\ref{sec:indexing} and returns the ones the answer should be built from. Its methods come largely from information retrieval, so the sparse, dense, and hybrid families are adopted here rather than invented, and structured retrieval searches the structures of Section~\ref{sec:topology} rather than flat chunks (Figure~\ref{fig:retrieval}).

\input{figures/fig_retrieval}

\subsection{Sparse retrieval}
\label{sec:ret-sparse}

Sparse retrieval, exemplified by BM25 \citep{bm25}, matches terms exactly and remains a
strong, cheap baseline: in a study of retrieval for language modelling, an off-the-shelf
BM25 retriever outperformed three popular dense retrievers \citep{incontextralm}.
Its weakness is paraphrase: it cannot match a query and a passage that share
meaning but not vocabulary. In practice it is retained as a complement to dense retrieval
rather than used alone, and it is often the safer choice for rare tokens such as identifiers,
proper names, and code snippets.

\subsection{Dense retrieval}
\label{sec:ret-dense}

Dense retrieval dominates modern RAG. DPR \citep{dpr} established the dual-encoder recipe:
embed query and passage separately, train with in-batch negatives so that positives score
higher than hard negatives, and retrieve by inner product. It is fast because the corpus is
embedded once, but the lack of query--document interaction limits its precision. ColBERT
\citep{colbert} restores interaction cheaply with late interaction: it keeps a vector per
token and scores a pair by the sum over query tokens of the maximum similarity to any
passage token. This is more expressive than a single-vector dot product and far cheaper than
a full cross-encoder. Multi-head RAG \citep{mrag} takes a different angle, retrieving with
per-attention-head representations and taking the top-$k$ per head, which helps queries that
have several distinct aspects. Augmentation-Adapted Retriever \citep{aar} adapts the retriever
to the generator so that the two cooperate as a plug-in.

\subsection{Hybrid retrieval}
\label{sec:ret-hybrid}

Hybrid retrieval runs a sparse and a dense retriever and fuses their rankings, typically by
weighted score combination or reciprocal rank fusion \citep{rrf}. It is a robust default: dense
retrieval covers paraphrase, sparse retrieval covers exact matches, and the two fail in
different places. Sparse--dense hybrids were designed precisely to capitalize on the
precision of sparse retrieval, and they sit at or near the top on large-scale retrieval
benchmarks \citep{luan2021sparse}; a systematic study of RAG design choices likewise makes
hybrid search its default retriever, combining BM25 with dense retrieval for strong
performance at relatively low latency \citep{wang2024searching}. Its cost is two indexes and
a fusion step, and its main design question is how to weight the two signals, which is usually
tuned on a validation set.

\subsection{Structured retrieval}
\label{sec:ret-structured}

Structured retrieval is the query-time counterpart of the structures built in
Section~\ref{sec:topology}: the index already holds a tree, a graph, or a hypergraph, and
retrieval walks it instead of scoring flat chunks. Tree traversal is usually top-down ---
descend layer by layer (RAPTOR \citep{raptor}), or
let the LM navigate (MemWalker \citep{memwalker}) --- while TreeRAG \cite{treerag},
DISRetrieval \cite{disretrieval}, and T-Retriever \cite{tretriever} walk a document, discourse, or graph-derived hierarchy to
collect the enclosing evidence. Graph retrieval
either spreads activation --- HippoRAG \citep{hipporag} runs Personalized PageRank from the query's entities
and returns the subgraph it activates --- or cuts a subgraph out, by
scoring triples (SubgraphRAG \citep{subgraphrag}) or beam-searching proposition paths
(PropRAG \citep{proprag}); GraphRAG \citep{graphrag} instead summarizes communities and answers by
map-reducing over them, while LightRAG \citep{lightrag} retrieves at two granularities,
from specific entities and relations to broader themes. Hypergraphs \citep{hypergraphrag} are
matched directly against entity and hyperedge embeddings. Where the
structure is a taxonomy, TaSR-RAG \citep{tasrrag} bridges the query and documents through it. PyRAG \citep{pyrag} decomposes a question into sub-questions and
writes executable code for each: retrieving, answering, and post-processing in a small
program.

\subsection{Discussion}
\label{sec:ret-disc}

The first three subsections ask a single question: which units score highest against this query. Their answers are all subsets of one index, so the outputs are the same kind of object and can be put side by side. The fourth asks what should be assembled for this query, and its answers need not be subsets at all: a summary of a region, a path of extracted propositions, a program that retrieves and then answers. A tradeoff exists: retrieval requiring LM calls adds inference latency and cost, while LM-free approaches may not be that adaptive.

%% file: figures/fig_retrieval.tex
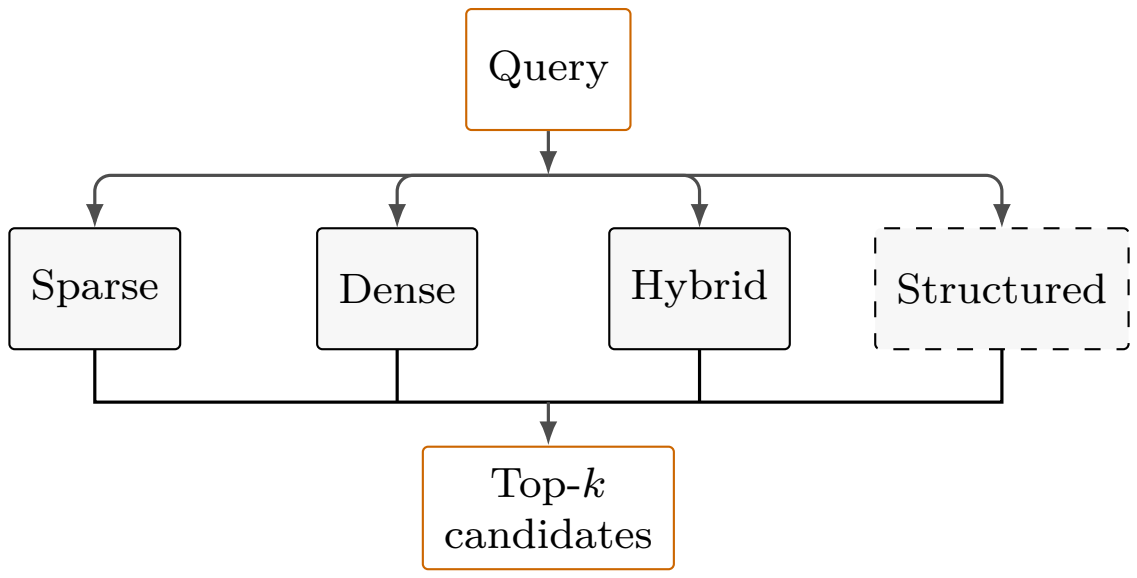
\begin{figure}[t]
\centering
\adjustbox{max width=\textwidth}{%
\begin{tikzpicture}[font=\scriptsize,
  d/.style={draw=BurntOrange, rounded corners=1pt, fill=white, minimum height=8mm, inner sep=4pt, align=center},
  s/.style={draw, rounded corners=1pt, fill=gray!6, minimum height=8mm, inner sep=4pt, align=center},
  o/.style={draw, dashed, rounded corners=1pt, fill=gray!6, minimum height=8mm, inner sep=4pt, align=center},
  a/.style={-{Latex[length=1.6mm]}, semithick, black!70},
  el/.style={rounded corners=3pt}
]
\node[d] (q)    at (3.0,1.45)  {Query};
\node[s] (sp)   at (0,0)       {Sparse};
\node[s] (de)   at (2.0,0)     {Dense};
\node[s] (hy)   at (4.0,0)     {Hybrid};
\node[o] (st)   at (6.0,0)     {Structured};
\node[d] (cand) at (3.0,-1.45) {Top-$k$\\candidates};
\draw[a] (q.south) -- (3.0,0.75);
\foreach \x in {0,2,4,6} { \draw[a,el] (3.0,0.75) -| (\x,0.4); }
\draw[semithick] (0,-0.4) -- (0,-0.75) -- (6,-0.75) -- (6,-0.4);
\draw[semithick] (2,-0.4) -- (2,-0.75);
\draw[semithick] (4,-0.4) -- (4,-0.75);
\draw[a] (3.0,-0.75) -- (cand.north);
\end{tikzpicture}%
}
\caption{Retrieval. Any one of these may feed the pipeline: sparse and dense score the
query against the index directly, hybrid fuses the two, and structured retrieval walks an explicit structure.
Structured retrieval is dashed because it requires an explicit structure built during indexing.}
\label{fig:retrieval}
\end{figure}

%% file: sections/post-retrieval.tex
\section{Post-retrieval Processing}
\label{sec:post-retrieval}

Retrieval returns candidates, not answers. Candidates can be noisy, redundant, and often too
long for the model to use well: the more context is injected, the more the model suffers
the lost-in-the-middle effect \cite{lost-in-the-middle}. Post-retrieval processing reranks,
refines, and finally fuses\footnote{We include fusion --- the choice of where the retrieved
evidence enters the model, which characterizes how generation is performed --- in
post-retrieval processing, because once the evidence has entered, generation itself is a
trivial language modeling process that does not merit a section of its own.} the retrieved
evidence into the generation process (Figure~\ref{fig:postretrieval}).

\input{figures/fig_postretrieval}

\subsection{Reranking}
\label{sec:reranking}

A first-stage retriever is optimized for recall; reranking reorders its output for precision. Normal reranking directly reranks based on a fixed budget, while dynamic reranking outputs different numbers of documents per query dynamically to be more adaptive.

\paragraph{Normal reranking.}
Early frameworks such as Re$^2$G \citep{re2g} simply inserted a reranker between retrieval
and generation. With LMs, reranking became a prompting
problem: RankGPT \citep{rankgpt} asks the model to rank a sliding window of candidates and
slides the window forward, listwise; Pairwise Ranking Prompting \citep{prp} instead compares
candidates in pairs. RankRAG \citep{rankrag}
unifies reranking and generation in a single model. CoRank \citep{corank} and SANE \citep{sane} perform LM-based reranking with compact features extracted from candidates for efficiency.

\paragraph{Dynamic reranking.}
ChunkRAG \citep{chunkrag} filters at
the chunk level. Sparse RAG \citep{sparserag} judges all candidates in parallel to determine whether each should be kept or dropped and reuses key-value cache during generation for efficiency. GenRT \citep{genrt} jointly models reranking and truncation, choosing how
many documents to keep as well as their order. DynamicRAG \citep{dynamicrag} uses the
generator's output as feedback to train a reranker supporting dynamic evidence budget, and EBCAR \citep{ebcar} reranks with
embedding-based context awareness to judge each candidate. The theme is the same: make the reranker adaptive to the
query and can output arbitrary numbers of evidence documents.

\subsection{Refinement}
\label{sec:refinement}

After reranking, the retrieved text may still be long and noisy, and much of it is actually redundant for the query. Refinement improves the content of the context by extraction, pruning, or restructuring before it is passed on.

\paragraph{Extraction and pruning.}
FilCo \citep{filco} learns to filter context, keeping sentences that actually help
generation. PRCA \citep{prca} treats the generator as a black-box and trains a reward-driven contextual adapter to extract helpful context. Token Elimination \citep{tokenelemination} removes tokens that the reader does not
need between decoder layers. 
RECOMP
\citep{recomp} compresses retrieved text into summaries, saving substantial context tokens. Provence \citep{provence} prunes context by training a sequence-to-sequence model jointly performing reranking and pruning using LM-labeled data.  Corrective RAG \citep{crag} adds a corrective step that evaluates retrieved
evidence and performs refinement and web search for plausible and poor knowledge, respectively. RAG-CSM \citep{RAG-CSM} trains a surrogate model to judge the contextual influence value for each evidence that guides pruning. These approaches directly prune noise and extract key information.

\paragraph{Query-time structuring.}
A newer line builds structure at inference rather than at index time, because index-time structures may not be sufficiently general to meet the requirements of all queries. StructRAG
\citep{structrag} selects a structure (graph, chunk, catalogue, algorithm, table) best suited to the
question and organizes the retrieved documents into it. DTCRS \citep{dtcrs} builds trees for recursive summarization of long inputs, and LongRefiner
\citep{longrefiner} refines documents based on hierarchical document trees. RAS \citep{ras}
iteratively retrieves and structures relevant text into a growing knowledge graph; SARG \citep{sarg} augments reasoning with a knowledge graph constructed based on retrieved context; and
SLIDERS \citep{sliders} organizes relevant long documents into relational databases to support complex reasoning. These methods
trade inference cost for the ability to adapt the structure to each query.

\subsection{Fusion}
\label{sec:fusion}

Fusion decides how evidence is combined with the model. The design space is
a ladder of increasingly invasive interventions: at the shallow end the evidence is
concatenated into the prompt and the model is left untouched, and at the deep end it is
written into the weights. The paragraphs below climb that ladder, passing through the
encoder and the output distribution on the way.

\paragraph{In-context level.}
Concatenating query and passages into the prompt is the default, studied systematically by
In-Context RALM \citep{incontextralm}; RA-DIT \citep{radit} tunes both retriever and
generator with instruction data so that they cooperate in this setting. No architectural
change is needed, so it works with any model that can take a longer prompt.

\paragraph{Encoder and token level.}
Fusion-in-Decoder \citep{fid} encodes the query with each passage separately and lets the
decoder attend to all of them, which scales to many passages; FiD-Light \citep{fidlight}
and COCOM \citep{cocom} make it more efficient with compact context embeddings. xRAG
\citep{xrag} compresses a passage into a single token via a learned projector, and REFRAG
\citep{refrag} adaptively expand chunk embeddings based on a learned policy. Retro \citep{retro} introduces chunked cross-attention for the interaction between query and document embeddings.

\paragraph{Decoding and distribution level.}
kNN-LM \citep{knnlm} interpolates the language model's next-token distribution with a
distribution read off the retrieved neighbours. Efficient
kNN-LM \citep{efficientknnlm} makes this practical by adaptive retrieval and more compact datastore and dimensions, and REST \citep{rest} turns retrieved
continuations into draft tokens for speculative decoding.
RePlug \citep{replug} ensembles each individual retrieved document at decoding time for black-box models. These
methods fuse at the level of output distributions, which requires access to logits.

\paragraph{Parametric level.}
The deepest option writes the evidence into the weights: P-RAG \citep{prag} trains
document-specific parameter deltas and merges the retrieved document's delta into the model
at inference, turning retrieval into a weight-space operation. This offers deep knowledge use
and avoids context-length limits, but it requires white-box access and enormous storage.

\subsection{Discussion}
\label{sec:post-disc}

The three stages differ in what they cost and when the cost is paid. Reranking and refinement are paid while answering, and their cost follows the size of what retrieval returned: one pass to order it, and another to organize it. Fusion is paid by building the system a certain way, committing the architecture or the weights before any query arrives. That decides how each can be used. The first two can be added to a system that already exists; the third decides what that system can be. Post-retrieval processing adds extra cost at inference time, which pays only when the final performance is improved.

%% file: figures/fig_postretrieval.tex
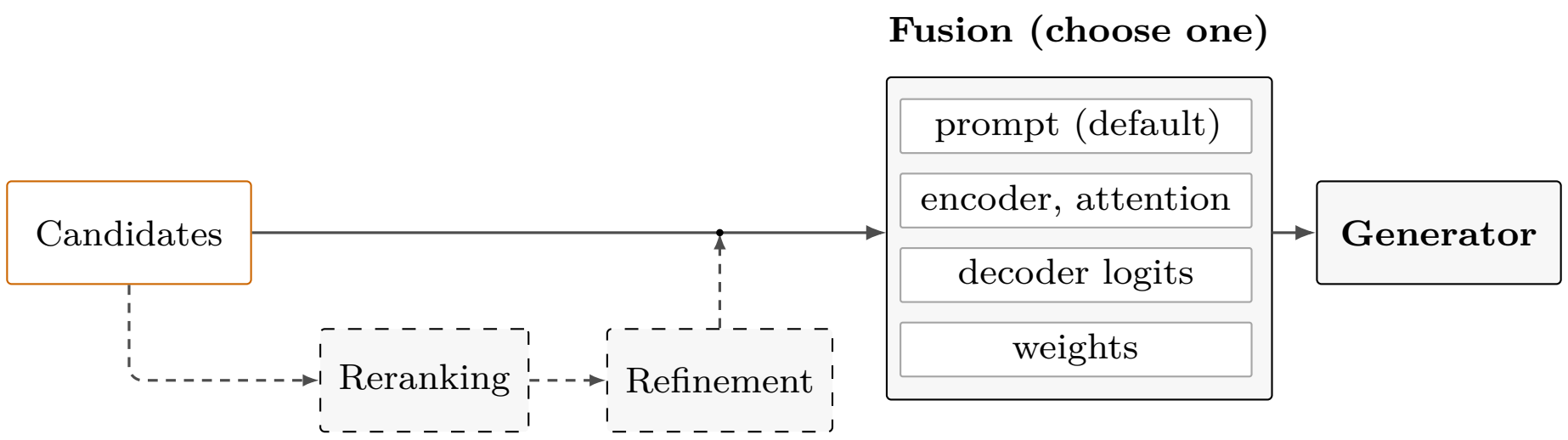
\begin{figure}[t]
\centering
\adjustbox{max width=\textwidth}{%
\begin{tikzpicture}[font=\scriptsize,
  d/.style={draw=BurntOrange, rounded corners=1pt, fill=white, minimum height=8mm, minimum width=1.9cm, inner sep=4pt, align=center},
  s/.style={draw, rounded corners=1pt, fill=gray!6, minimum height=8mm, inner sep=4pt, align=center},
  o/.style={draw, dashed, rounded corners=1pt, fill=gray!6, minimum height=8mm, inner sep=4pt, align=center},
  rung/.style={draw=black!35, rounded corners=0.5pt, fill=white, text width=2.65cm, align=center,
               minimum height=4.2mm, inner sep=1.2pt},
  a/.style={-{Latex[length=1.6mm]}, semithick, black!70},
  oa/.style={-{Latex[length=1.3mm]}, semithick, dashed, black!70, dash pattern=on 2.2pt off 1.8pt},
  el/.style={rounded corners=3pt}
]
\draw[rounded corners=1pt, fill=gray!6] (5.9,0.91) rectangle (8.9,-1.60);
\node[font=\scriptsize\bfseries, anchor=south, yshift=2pt] at (7.4,0.91) {Fusion (choose one)};
\node[rung, anchor=west] (f1) at (6.0,0.53)   {prompt (default)};
\node[rung, anchor=west] (f2) at (6.0,-0.05)  {encoder, attention};
\node[rung, anchor=west] (f3) at (6.0,-0.63)  {decoder logits};
\node[rung, anchor=west] (f4) at (6.0,-1.21)  {weights};
\node[s, minimum width=1.9cm, font=\scriptsize\bfseries] (gen) at (10.2,-0.30) {Generator};
\node[d] (cand) at (0,-0.30) {Candidates};
\draw[a] (cand.east) -- (5.9,-0.30);
\node[o] (rerank) at (2.3,-1.45) {Reranking};
\node[o] (refine) at (4.6,-1.45) {Refinement};
\draw[oa,el] (cand.south) |- (rerank.west);
\draw[oa] (rerank.east) -- (refine.west);
\draw[oa] (refine.north) -- (4.6,-0.30);
\fill (4.6,-0.30) circle (0.8pt);
\draw[a] (8.9,-0.30) -- (gen.west);
\end{tikzpicture}%
}
\caption{Post-retrieval. Candidates may go straight on (solid), or be reranked and refined
first (dashed). Fusion is the interface at which the
evidence enters the model, and a system picks exactly one: the prompt (the default), the
encoder, the decoder's logits, or the weights.}
\label{fig:postretrieval}
\end{figure}

%% file: sections/orchestration.tex
\section{Orchestration and Workflow}
\label{sec:orchestration}

The previous stages describe what each step does; orchestration describes
how they are sequenced. We distinguish five classes, ordered by how much freedom the
workflow has (Figure~\ref{fig:orch}): \emph{linear}, a single fixed pass; \emph{branching},
parallel paths whose results are merged; \emph{iterative}, retrieval and generation
interleaved over rounds; \emph{recursive}, a query decomposed and its parts handled in turn;
and \emph{agentic}, in which the model itself chooses the workflow. The first four keep the
control flow fixed in advance and share a common vocabulary, so we treat them here. Agentic
RAG breaks with that vocabulary --- the path is inferred at run time rather than designed ---
and it has grown into a paradigm of its own; we therefore give it a dedicated Section~\ref{sec:agentic}.

\input{figures/fig_orchestration}

\subsection{Linear}
\label{sec:orch-linear}

The linear workflow runs each stage once, in a fixed order: the retrieve-then-read
baseline. RAG \citep{lewis2020rag} is its canonical instance: a dense retriever feeds
passages to a sequence-to-sequence generator, and Re2G \citep{re2g} slots a reranker
between retrieval and generation. The chain can run longer: MedRGAG \citep{medrgag} uses
three sequential stages for medical question answering, retrieving from several medical
corpora, generating the background documents that cover the knowledge gaps the retrieval
left, and then selecting a final evidence set from the retrieved and generated documents
together. Most early methods are linear, and it remains the right default when queries are
simple and the corpus is well indexed. Its limitation is that a single
retrieval cannot recover from a bad first query, and a single generation cannot request
more evidence.

\subsection{Branching}
\label{sec:orch-branch}

Branching splits the workflow into parallel paths and merges their results. GRG
\citep{grg} uses a generator--retriever--generator pattern: the generator receives both model-generated and retrieved documents based on the query. BlendFilter \citep{blendfilter} blends knowledge retrieved based on queries augmented by both external and internal knowledge.
RePlug \citep{replug} sends each retrieved document into the model on its own pass and
ensembles the resulting output distributions, trading compute for accuracy. Branching is
attractive when a query can be attacked from several angles at once, at the cost of more
inference per query.

\subsection{Iterative}
\label{sec:orch-iter}

Iterative workflows interleave retrieval and generation. IRCoT \citep{ircot} retrieves, then
generates a chain of thought whose content becomes the next retrieval query, and repeats.
Iter-RetGen \citep{iterretgen} uses the query augmented by the generated answer as the next retrieval key. FLARE
\citep{flare} drives iteration by generation uncertainty: it generates sentence by
sentence and retrieves when the model is unsure of the next sentence. Self-Memory
\citep{selfmem} lifts generation by retrieving from its own past outputs. RAS \citep{ras}
interleaves retrieval planning with incremental construction of a question-specific
knowledge graph, so what grows across rounds is the evidence structure itself. ReflectiveRAG
\citep{reflectiverag} deploys a small model to judge whether the retrieval is sufficient and refine the query for the next retrieval if not. Iteration is the simplest way to handle multi-step
questions, and its main cost is latency: each round is a full retrieve-and-generate pass. This cost is mitigated by Q-RAG \cite{Q-RAG}, which performs multi-turn generation-free retrieval with a tuned embedder.

\subsection{Recursive}
\label{sec:orch-rec}

Recursive workflows decompose a query and recurse on the parts. Tree of Clarifications
\citep{toc} recursively decomposes an ambiguous question into interpretations until each is
answerable, then aggregates. RA-ISF \citep{raisf} decomposes a query if it can be answered by neither the model's own knowledge nor retrieved documents. TreeRare
\citep{treerare} builds a syntax tree for the query and uses subtrees as retrieval keys.
Recursion suits questions that are genuinely composite.

\subsection{Discussion}
\label{sec:orch-disc}

The spectrum trades cost for adaptivity: linear is cheapest and the most predictable, and each added degree of freedom buys robustness on harder questions at a cost in latency and tokens. The classes differ, though, in whether anything decides that another step should be taken. Linear fixes the sequence before the query arrives and branching runs every path it was given, so neither does. Iteration and recursion have to, because a loop needs a rule for going round again, and that rule is a component that can be wrong in a way a fixed sequence cannot: it can stop a round that was needed. Such a rule answers only whether to continue. Agentic RAG hands the same control to the model, which then decides what the next step is as well; Section~\ref{sec:agentic} follows that change through.

%% file: figures/fig_orchestration.tex
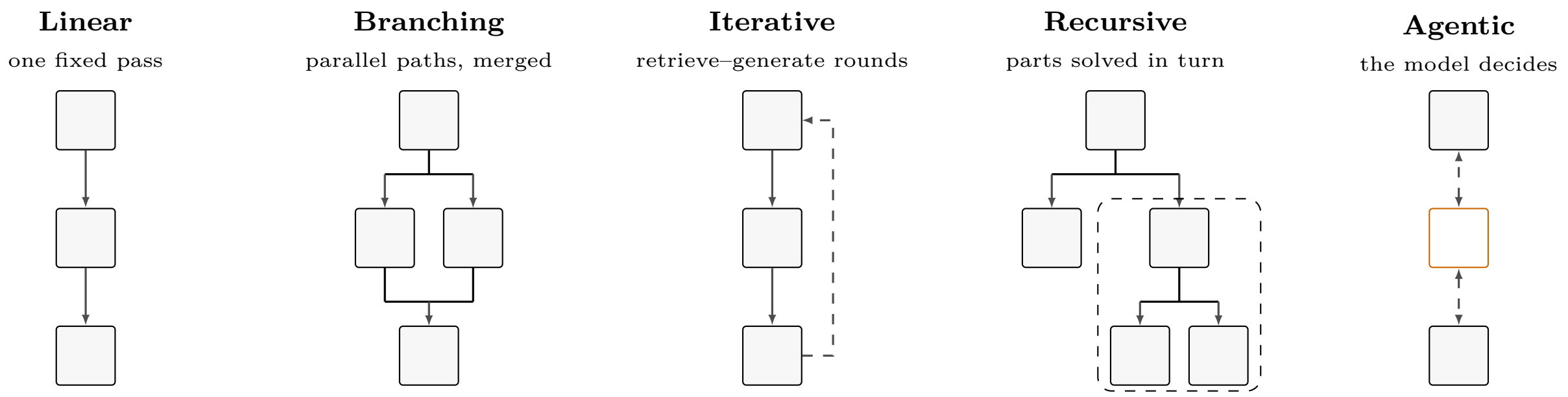
\begin{figure}[t]
\centering
\adjustbox{max width=\textwidth}{%
\begin{tikzpicture}[font=\scriptsize,
  b/.style={draw, rounded corners=1pt, fill=gray!6, minimum size=6mm, inner sep=0pt},
  t/.style={draw=BurntOrange, rounded corners=1pt, fill=white, minimum size=6mm, inner sep=0pt},
  a/.style={-{Latex[length=1.1mm]}, semithick, black!70},
  da/.style={-{Latex[length=1.1mm]}, semithick, dashed, black!70},
  dab/.style={{Latex[length=1.1mm]}-{Latex[length=1.1mm]}, semithick, dashed, black!70},
  el/.style={rounded corners=2pt},
  hd/.style={anchor=south, align=center, inner sep=0pt}
]
\node[hd] at (0,0)    {{\bfseries Linear}\\[1pt]{\tiny\mdseries one fixed pass}};
\node[hd] at (3.5,0)  {{\bfseries Branching}\\[1pt]{\tiny\mdseries parallel paths, merged}};
\node[hd] at (7.0,0)  {{\bfseries Iterative}\\[1pt]{\tiny\mdseries retrieve--generate rounds}};
\node[hd] at (10.5,0) {{\bfseries Recursive}\\[1pt]{\tiny\mdseries parts solved in turn}};
\node[hd] at (14.0,0) {{\bfseries Agentic}\\[1pt]{\tiny\mdseries the model decides}};

\node[b] (l1) at (0,-0.50) {};
\node[b] (l2) at (0,-1.70) {};
\node[b] (l3) at (0,-2.90) {};
\draw[a] (l1)--(l2); \draw[a] (l2)--(l3);

\node[b] (b1) at (3.50,-0.50) {};
\node[b] (b2) at (3.05,-1.70) {};
\node[b] (b3) at (3.95,-1.70) {};
\node[b] (b4) at (3.50,-2.90) {};
\draw[semithick] (b1.south) -- (3.50,-1.05);
\draw[semithick] (3.05,-1.05) -- (3.95,-1.05);
\draw[a] (3.05,-1.05) -- (3.05,-1.40);
\draw[a] (3.95,-1.05) -- (3.95,-1.40);
\draw[semithick] (3.05,-2.00) -- (3.05,-2.35);
\draw[semithick] (3.95,-2.00) -- (3.95,-2.35);
\draw[semithick] (3.05,-2.35) -- (3.95,-2.35);
\draw[a] (3.50,-2.35) -- (3.50,-2.60);

\node[b] (i1) at (7.0,-0.50) {};
\node[b] (i2) at (7.0,-1.70) {};
\node[b] (i3) at (7.0,-2.90) {};
\draw[a] (i1)--(i2); \draw[a] (i2)--(i3);
\draw[da] (i3.east) -- (7.62,-2.90) -- (7.62,-0.50) -- (7.30,-0.50);

\node[b] (r1) at (10.50,-0.50) {};
\node[b] (r2) at (9.85,-1.70) {};
\node[b] (r3) at (11.15,-1.70) {};
\node[b] (r4) at (10.75,-2.90) {};
\node[b] (r5) at (11.55,-2.90) {};
\draw[semithick] (r1.south) -- (10.50,-1.05);
\draw[semithick] (9.85,-1.05) -- (11.15,-1.05);
\draw[a] (9.85,-1.05) -- (9.85,-1.40);
\draw[a] (11.15,-1.05) -- (11.15,-1.40);
\draw[semithick] (11.15,-2.00) -- (11.15,-2.35);
\draw[semithick] (10.75,-2.35) -- (11.55,-2.35);
\draw[a] (10.75,-2.35) -- (10.75,-2.60);
\draw[a] (11.55,-2.35) -- (11.55,-2.60);
\draw[dashed, rounded corners=4pt] (10.32,-3.26) rectangle (11.98,-1.30);

\node[b] (ga) at (14.0,-0.50) {};
\node[t] (gm) at (14.0,-1.70) {};
\node[b] (gb) at (14.0,-2.90) {};
\draw[dab] (gm)--(ga); \draw[dab] (gm)--(gb);
\end{tikzpicture}%
}
\caption{The orchestration spectrum.
Orange marks the model itself and dashed arrows mark run-time, adaptive control; in the
recursive column the dashed box marks one part being decomposed again by the same
procedure.}
\label{fig:orch}
\end{figure}

%% file: sections/agentic.tex
\section{Agentic RAG}
\label{sec:agentic}

In Agentic RAG, the LM itself decides the control flow of retrieval at inference time (Figure~\ref{fig:agentic-workflow}). Based on its current reasoning state and retrieved evidence, the model can decide whether to retrieve, how to formulate a query, which information source to use, and when to stop. What matters is who makes these decisions, not whether the system involves multiple retrieval rounds, multiple components, or external tools. Recent work extends this control in several directions.

\input{figures/fig_agentic}

\subsection{Self-Reflective retrieval}
\label{sec:agentic-reflection}

An early step toward agentic retrieval is to make retrieval conditional. Self-reflective methods let the model decide whether external evidence is needed and evaluate what comes back, instead of retrieving whenever the pipeline reaches a predefined stage. Self-RAG \citep{selfrag} illustrates this shift. It uses reflection tokens to support selective retrieval and to assess both retrieved passages and generated responses. Part of retrieval control thus moves from a fixed pipeline to the model itself. From an agentic perspective, what matters is not the reflection mechanism itself but the two capabilities it introduces, selective retrieval and self-evaluation. Later systems take this further, extending model control from whether to retrieve to how the retrieval process should proceed.

\subsection{Controlling the retrieval trajectory}
\label{sec:agentic-planning}

Beyond deciding whether retrieval is needed, an agent can also control how the retrieval process unfolds. The key change is from executing a predetermined sequence of retrieval steps to adjusting the search trajectory based on intermediate reasoning and newly observed evidence. Search-o1 \citep{searcho1} searches on demand during reasoning, and CoRAG \citep{CoRAG} and DeepRAG \citep{deeprag} decide the next retrieval step based on earlier results or the current reasoning state. Other methods move from sequential search to structured navigation. A-RAG \citep{arag}, DocNavRAG \citep{docnavrag}, and VecTree-RAG \citep{vectreerag} organize retrieval over hierarchical, document-level, or tree-structured information, while Graph-R1 \citep{graphr1} and LogicRAG \citep{LogicRAG} let retrieval follow graph structures or dependencies between subqueries. Iteration or decomposition alone does not make retrieval agentic, since both can be implemented by a fixed pipeline. The real difference is whether the search path can be adjusted during inference by the model based on the information obtained so far.

\subsection{Multi-Agent retrieval}
\label{sec:agentic-multi}

Retrieval control can also be distributed across multiple LM agents with specialized roles. Existing methods differ mainly in how this specialization is organized. MA-RAG \citep{marag} separates planning, step definition, evidence extraction, and answer generation; JADE \citep{jade} uses multiple specialized agent roles coordinated by a central planner; and Search-R2 \citep{searchr2} splits search-integrated reasoning into actor and refiner roles. This flexibility brings coordination costs, mainly from extra communication and duplicated evidence, so whether multi-agent design is worth it depends on whether specialization gains exceed these costs.

\subsection{Learning agentic retrieval policies}
\label{sec:agentic-learning}

Recent work does not only expand which parts of retrieval are under model control. It also asks how that control is acquired. The main change is from specifying retrieval behavior through prompts or hand-written rules to learning a retrieval policy from interaction. Search-R1 \citep{searchr1} uses reinforcement learning to teach the model to interleave reasoning and search, and DeepRAG \citep{deeprag} learns when external retrieval is better than internal reasoning. Learned retrieval control also extends to different retrieval mechanisms. Graph-R1 \citep{graphr1} learns policies for interacting with graph-based retrieval, and MMOA-RAG \citep{MMOA-RAG} jointly optimizes cooperating agents across query rewriting, retrieval, selection, and generation. In these systems, control that was once specified by hand is increasingly learned from interaction. This shift makes credit assignment and retrieval cost central problems. If rewards are computed only from the final answer, they cannot show which intermediate searches were actually useful, so finer supervision is needed, such as the hierarchical process rewards in HiPRAG \citep{hiprag}, the information-gain-based turn-level rewards in IGPO \citep{igpo}, and the fine-grained, efficiency-aware rewards in TeaRAG \citep{TeaRAG}. Learning an effective retrieval policy is not just about learning to search per se. The policy has to judge which retrieval actions provide useful information and which calls are unnecessary. It therefore has to weigh information gain against retrieval cost and decide when further retrieval is no longer worth it.

\subsection{Discussion}
\label{sec:agentic-disc}

What defines Agentic RAG is who decides the retrieval control flow. The extent of agentic behavior depends on how much control the model has, from individual retrieval decisions to the entire search trajectory. This model-directed control also connects Agentic RAG to recursive self-improvement (RSI). Both use feedback from previous actions to guide later decisions. In Agentic RAG, retrieved evidence updates the model's reasoning state, which determines whether to retrieve again and when to stop. More control, however, is not always better. Dynamic retrieval is useful when the retrieval path cannot be specified in advance, while a fixed workflow may be cheaper and more reliable for simpler queries. Agentic systems therefore need to be evaluated on retrieval cost, latency, stopping behavior, and evidence quality, not just final-answer quality. The hard part is giving the model enough flexibility to search well without paying for searches that do not help.

%% file: figures/fig_agentic.tex
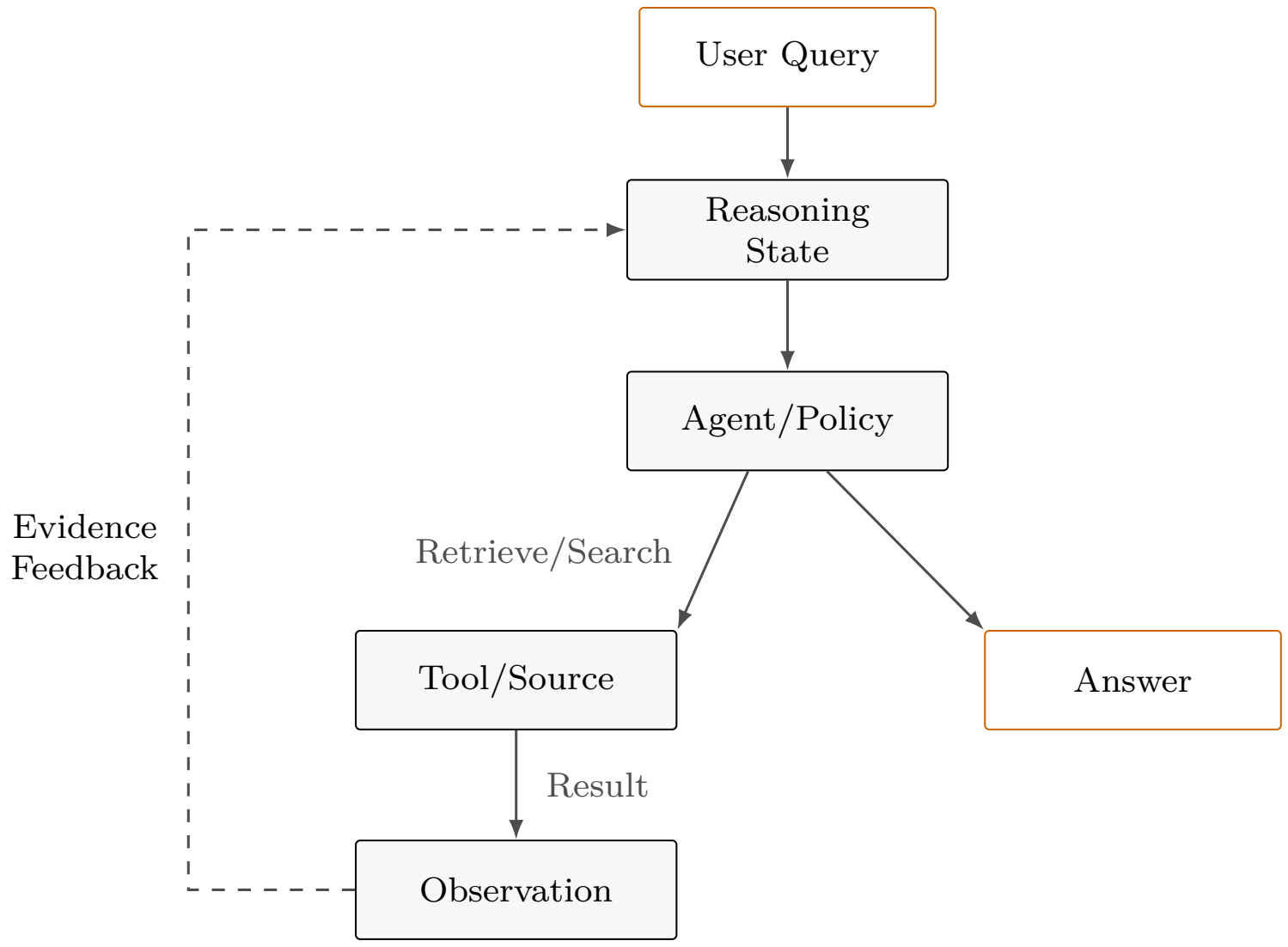
\begin{figure}[t]
\centering
\adjustbox{max width=0.60\textwidth}{%
\begin{tikzpicture}[font=\scriptsize,
  d/.style={
    draw=BurntOrange,
    rounded corners=1pt,
    fill=white,
    minimum height=8mm,
    minimum width=24mm,
    inner sep=4pt,
    align=center
  },
  s/.style={
    draw,
    rounded corners=1pt,
    fill=gray!6,
    minimum height=8mm,
    minimum width=26mm,
    inner sep=4pt,
    align=center
  },
  a/.style={
    -{Latex[length=1.6mm]},
    semithick,
    black!70
  },
  oa/.style={
    -{Latex[length=1.6mm]},
    semithick,
    dashed,
    black!70
  }
]

\node[d] (query) at (0,4.2)
{User Query};

\node[s] (state) at (0,2.8)
{Reasoning\\State};

\node[s] (agent) at (0,1.25)
{Agent/Policy};

\node[s] (tool) at (-2.2,-0.85)
{Tool/Source};

\node[d] (answer) at (2.8,-0.85)
{Answer};

\node[s] (obs) at (-2.2,-2.55)
{Observation};

\draw[a]
(query.south) -- (state.north);

\draw[a]
(state.south) -- (agent.north);

\draw[a]
($(agent.south)+(-0.32,0)$)
--
node[
  pos=0.58,
  left=4pt,
  yshift=2pt,
  font=\scriptsize
]
{Retrieve/Search}
(tool.north east);

\draw[a]
($(agent.south)+(0.32,0)$)
--
(answer.north west);

\draw[a]
(tool.south)
--
node[
  midway,
  right=3pt,
  font=\scriptsize
]
{Result}
(obs.north);

\coordinate (feedback-left)
  at ($(obs.west)+(-1.35,0)$);

\coordinate (feedback-top)
  at (feedback-left |- state.west);

\draw[oa]
(obs.west)
-- (feedback-left)
-- (feedback-top)
-- (state.west);

\node[
  font=\scriptsize,
  align=center,
  left=3pt
]
at ($(feedback-left)!0.52!(feedback-top)$)
{Evidence\\Feedback};

\end{tikzpicture}%
}

\caption{A typical agentic retrieval loop. The agent selects
actions from its current reasoning state, interacts with external tools or
information sources, and updates its state from the resulting observations.}
\label{fig:agentic-workflow}
\end{figure}

%% file: sections/evaluation.tex
\section{Evaluation}
\label{sec:eval}

Evaluation requires four choices: the task, the basis for judging results, the evaluation procedure, and the metrics to report. RAG evaluation examines retrieval and generation. Retrieval evaluation asks whether the system obtains the evidence, while generation evaluation asks whether the answer is correct and supported~\citep{ragchecker}. Agentic RAG also requires accounting for the number and cost of search calls~\citep{browsecompplus}.

\subsection{A framework for evaluation}
\label{sec:eval-data}
\label{sec:eval-frameworks}
\label{sec:eval-metrics}

Each of these four choices shapes what an evaluation can establish. First, the sources of the questions and evidence distinguish open-domain QA~\citep{naturalquestions,triviaqa}, multi-hop QA~\citep{hotpotqa,twowikimultihopqa}, and long-document QA~\citep{narrativeqa,longbench}. Second, annotations and reference materials determine which errors can be detected. Answer-only labels support correctness judgments, while annotations of evidence or claims enable checks of whether answers are supported, as illustrated by the supporting facts in HotpotQA~\citep{hotpotqa} and the citations in ALCE~\citep{alce}. Third, the evaluation procedure may measure retrieval and generation separately~\citep{beir} or score the system end to end~\citep{hotpotqa}. Fourth, common metrics include exact match (EM) and F1 for short answers~\citep{naturalquestions}, ROUGE for long answers~\citep{narrativeqa}, and nDCG for retrieval~\citep{beir}. Faithfulness and citation quality call for model judgments or human verification~\citep{ragas,alce}. Table~\ref{tab:eval-resources} compares representative resources along these dimensions.

In practice, the most widely used settings are short-answer open-domain and multi-hop QA with Wikipedia as a retrieval corpus~\citep{hotpotqa,naturalquestions,triviaqa,twowikimultihopqa,musique}. String matching, fixed corpora within a shared experimental setting, and public baselines make results easy to compare. When evaluation uses final-answer correctness alone, it conflates evidence that was retrieved but ignored with evidence that was never retrieved~\citep{ragchecker}. Standard setups of this kind also leave variation in context length, domain, and information freshness outside the test conditions~\citep{longbench,rageval,cragbenchmark}.

Other resources address these gaps. Long-document and domain benchmarks vary how evidence is distributed~\citep{qasper,narrativeqa,quality,longbench}. Controlled resources vary the evidence supplied to expose sensitivity to noise, abstention behavior, and reasoning across documents~\citep{rgb,cragbenchmark}. Calibration and diagnostic frameworks change the granularity of assessment to help locate errors in retrieval or generation~\citep{ragchecker,ares}. Toolkits standardize the implementation to make experiments reproducible~\citep{bergen,flashrag}. These resources remain less common in routine reporting. Diagnostic frameworks score claims, spans, or citations~\citep{ragchecker,alce}, automatic metrics can rely on model judges~\citep{ragas}, and calibrated judgments can incorporate human labels~\citep{ares}. The resulting scores are difficult to align with established baselines, so routine benchmark reporting and diagnostic evaluation coexist as distinct practices.

\input{tables/eval-resources}

\subsection{Evaluating agentic retrieval}
\label{sec:eval-agentic}

Agentic RAG still primarily uses benchmarks inherited from non-agentic systems. Open-domain and multi-hop QA remain central~\citep{hotpotqa,naturalquestions,triviaqa,twowikimultihopqa,musique}. Evaluation mainly uses end-to-end metrics; some studies also report token cost and latency~\citep{rise}. Final-answer scores leave the search process unobserved, including how often the system searched, when it stopped, and which steps were necessary. Under such evaluation, ablations can provide indirect evidence about search trajectories.

Three designs address this gap. AgenticRAGTracer~\citep{agenticragtracer} uses questions with hop-level annotations and logical decompositions to diagnose individual steps while allowing multiple valid action sequences. BrowseComp-Plus~\citep{browsecompplus} fixes the corpus and supporting documents, enabling separate comparisons of retriever capability and agent policy. DeepResearch Bench~\citep{deepresearchbench} uses expert-authored tasks and RACE and FACT to assess final reports. The designs specify, respectively, the question structure, the evidence supply, and the criteria for judging the output. However, these approaches have not yet been widely applied and appear mainly in evaluation research.


\subsection{Discussion}
\label{sec:eval-disc}
\label{sec:eval-leakage}

A RAG score admits two explanations: the model already possessed the relevant knowledge, or its answer depended on the supplied evidence. Public test material may overlap with training data, and the RETRO analysis~\citep{retro} shows how filtering out overlapping material changes the interpretation of scores. Comparing closed-book answering, answering with retrieved context, and answering with gold evidence helps distinguish the two explanations, as in FRAMES~\citep{frames}. A correct answer given gold evidence can still draw on prior knowledge. A citation may simply supply a source for an answer the model already knew, so citation accuracy alone cannot establish that the cited passage caused the response~\citep{alce,countermem}. CounterMem~\citep{countermem} holds the system configuration fixed and removes retrieved evidence or supplies contradictory evidence to observe directly whether the answer changes. Attributing a gain in answer correctness to retrieval on a given question therefore requires showing that the model could not answer correctly without the evidence and that the supplied evidence changed its answer. Without either condition, the same score remains open to a different explanation. Meanwhile, even on the same dataset, results can vary with the task subset, corpus, passage boundaries, or scoring procedure~\citep{kilt,beir,longbench}. Agentic evaluations also differ in the search paths and resource budgets available~\citep{browsecompplus,rise,agenticragtracer}. Comparing gains from external knowledge thus requires recording which evidence was available to the system, whether its answer depended on that evidence, and the cost of retrieving it~\citep{browsecompplus,frames,countermem}.

%% file: tables/eval-resources.tex
\begingroup
\fontsize{8.2}{9.2}\selectfont
\setlength{\tabcolsep}{2.5pt}
\renewcommand{\arraystretch}{1.08}
\setlength{\LTcapwidth}{1.00\linewidth}
\newlength{\evaltablecontentwidth}
\setlength{\evaltablecontentwidth}{\dimexpr1.00\linewidth-8\tabcolsep\relax}
\setlength{\LTleft}{\fill}
\setlength{\LTright}{\fill}
\begin{longtable}{@{}>{\raggedright\arraybackslash}p{0.21\evaltablecontentwidth} >{\raggedright\arraybackslash}p{0.23\evaltablecontentwidth} >{\raggedright\arraybackslash}p{0.22\evaltablecontentwidth} >{\raggedright\arraybackslash}p{0.17\evaltablecontentwidth} >{\raggedright\arraybackslash}p{0.17\evaltablecontentwidth}@{}}
\caption{Data sources and measurement procedures used in RAG evaluation. Task answers, provenance, and claim- or span-level annotations make different parts of the retrieval--generation process observable.}
\label{tab:eval-resources}\\
\toprule
\textbf{Resource} & \textbf{Data} & \textbf{Annotations and references} & \textbf{Procedure} & \textbf{Metrics} \\
\midrule
\endfirsthead
\multicolumn{5}{@{}l}{\textbf{Table \thetable\ continued}}\\
\toprule
\textbf{Resource} & \textbf{Data} & \textbf{Annotations and references} & \textbf{Procedure} & \textbf{Metrics} \\
\midrule
\endhead
\midrule
\multicolumn{5}{r@{}}{Continued on next page}\\
\endfoot
\bottomrule
\endlastfoot
\multicolumn{5}{@{}l}{\textit{Test corpora and labeled tasks}}\\*[3pt]
Natural Questions~\citep{naturalquestions}, TriviaQA~\citep{triviaqa}, PopQA~\citep{eval_popqa}, WebQuestions~\citep{eval_webquestions}, SQuAD~\citep{eval_squad} & Search queries with Wikipedia; trivia questions; entity-centric long-tail questions; Freebase fact questions; Wikipedia passages with answer spans & Human short-answer annotations, with evidence documents varying by resource; PopQA questions are template-generated; SQuAD annotates answer spans & Retrieve, then answer, in an open-domain or open-book setting & EM, F1, accuracy \\
\addlinespace[1.5pt]
HotpotQA~\citep{hotpotqa}, 2WikiMultiHopQA~\citep{twowikimultihopqa}, MuSiQue~\citep{musique}, Bamboogle~\citep{eval_bamboogle}, StrategyQA~\citep{eval_strategyqa} & Wikipedia passages with linked or composed multi-hop questions; human-written two-hop questions; yes/no questions requiring implicit reasoning & Human supporting facts in HotpotQA; structured construction or manual decomposition in the others & Score final answers and, where annotated, supporting evidence & Answer EM/F1; evidence and joint scores \\
\addlinespace[1.5pt]
NarrativeQA~\citep{narrativeqa}, QASPER~\citep{qasper}, QuALITY~\citep{quality}, LongBench~\citep{longbench}, RAGEval~\citep{rageval} & Books and scripts; research papers; long articles; English/Chinese long-context tasks; generated finance, law and medicine scenarios & Human questions with free-form answers; evidence spans; options and validation; task labels; LLM generation with human quality checks & Evaluate summary versus full-story reading, paper QA, long-passage choice, and task performance & BLEU/ROUGE-L; answer and evidence F1; accuracy; completeness and hallucination \\
\addlinespace[1.5pt]
BEIR~\citep{beir}, MS MARCO~\citep{eval_msmarco} & Heterogeneous query-corpus collections; real queries with web passages & Dataset-specific document relevance judgments; query logs and human relevance judgments & Document ranking and large-scale passage retrieval & nDCG@10, MRR@10, recall \\
\addlinespace[1.5pt]
KILT~\citep{kilt} & Knowledge-intensive tasks aligned to one Wikipedia snapshot & Task answers and provenance mapped to the shared corpus & Evaluate task performance and evidence provenance together & Task scores, provenance retrieval, KILT score \\
\addlinespace[1.5pt]
RGB~\citep{rgb}, CRAG~\citep{cragbenchmark} & English/Chinese QA with controlled context; factual questions with web and structured resources & LLM-generated QA with manual verification and controlled evidence edits; expert review & Fix the question and vary the evidence; controlled resource access with abstention & Accuracy and rejection by condition; abstention-aware score \\
\addlinespace[1.5pt]
FEVER~\citep{eval_fever} & Claims rewritten from Wikipedia sentences & Human veracity labels with supporting evidence & Verify claims against Wikipedia & Label accuracy, evidence retrieval \\
\addlinespace[1.5pt]
\multicolumn{5}{@{}l}{\textit{Scoring, diagnosis, and experimental procedures}}\\*[3pt]
RAGAs~\citep{ragas} & User-supplied questions, retrieved contexts and answers & Basic metrics need no gold answer & LLM or embedding-based scoring & Faithfulness, answer relevance, correctness \\
\addlinespace[1.5pt]
ARES~\citep{ares}, RAGChecker~\citep{ragchecker} & Target-domain passages with synthetic samples; adapted public QA or RAG traces & Synthetic labels with human calibration; reference answers, claims and entailment checks & Train judges and correct aggregate estimates; decompose answers to diagnose retrieval and generation errors & Relevance, faithfulness, confidence intervals; claim P/R/F1 and context utilization \\
\addlinespace[1.5pt]
ALCE~\citep{alce} (with ASQA~\citep{eval_asqa}) & ASQA, QAMPARI and ELI5 with retrieval corpora & References and generated citations, with human checks for meta-evaluation & Score answer quality and whether cited passages support claims & Correctness, coverage, citation precision and recall \\
\addlinespace[1.5pt]
BERGEN~\citep{bergen}, FlashRAG~\citep{flashrag} & Configurable corpora, task data and system components & Use the selected task's labels & Run reproducible baselines and component combinations & Task metrics set by the experimental configuration \\
\addlinespace[1.5pt]
\end{longtable}
\endgroup

%% file: sections/applications.tex
\section{Applications}
\label{sec:apps}

RAG has been widely applied in a wide range of tasks and domains. We hereby discuss representative applications of RAG, including general question answering and assistance, domain-specific question answering, summarization and synthesis, creation and ideation, fact verification and text revision, and code completion.


\paragraph{General question answering and personalized assistance.}
In general QA, retrieval supplies evidence for the answer. ChatQA~\citep{app_liu2024chatqa}
pairs a conversational dense retriever with an instruction-tuned generator.
In personalized assistance, user-specific context can help interpret the request or justify
the response.
The retrieval baselines in LaMP-QA~\citep{app_salemi2025lampqa} use past questions and their
descriptions to infer the user's preferences for long-form answers. G-Refer~\citep{app_li2025grefer}
draws on paths and textual attributes in a user--item graph to explain recommendations.

\paragraph{Domain-specific question answering.}
Domain QA can fail when the relevant knowledge is missing in the generator LM's training.
The MedRAG~\citep{app_xiong2024medrag} toolkit compares choices of corpus, retriever, and
generator in medical QA. FinGEAR~\citep{app_li2025fingear} uses financial terminology and
filing hierarchies to locate relevant passages. Once passages are in context, the problem
shifts to evidence use. The pipeline evaluated on JustifIT-QA~\citep{app_dallanoce2026justifit}
uses judge feedback and the same retrieved provisions to revise a legal answer.
RAFT~\citep{app_zhang2024raft} addresses evidence use at training time by teaching the
generator to identify supporting passages among distractors.

\paragraph{Summarization and research synthesis.}
Long-form synthesis needs evidence for several parts of the output. Two design choices are
how to organize this evidence and when to collect it. GoR~\citep{app_zhang2025gor} links document
chunks to earlier model responses in a graph for summarization.
STORM~\citep{app_shao2024storm} collects sources from several perspectives before outlining
to cover the topic broadly. Retrieval during writing can target specific gaps:
OpenScholar~\citep{app_asai2026openscholar} uses feedback on a cited draft to decide whether
to revise it or seek more evidence; WebThinker~\citep{app_li2025webthinker} interleaves web
search with drafting.

\paragraph{Content creation and scientific ideation.}
Retrieval gives open-ended generation a starting point. RACo~\citep{app_yu2022raco} uses
commonsense text to connect a given set of concepts in a plausible sentence. In scientific
ideation, prior work can also help screen the output: SciMON~\citep{app_wang2024scimon} draws
on the literature to propose ideas and revises proposals that resemble earlier work too closely.

\paragraph{Fact verification and text revision.}
Verification and revision both retrieve evidence for claims in existing text.
FIRE~\citep{app_xie2025fire} checks whether the evidence is sufficient for a verdict and
searches again if needed. RARR~\citep{app_gao2023rarr} turns a passage into search questions
and uses the results to edit unsupported claims. Evidence thus gives verification a stopping
rule and helps revision preserve supported text.

\paragraph{Repository-level code completion.}
A code prefix may omit identifiers that point to relevant files elsewhere in the repository.
RepoCoder~\citep{app_zhang2023repocoder} uses an initial completion to expand the query for
the next retrieval pass. Repository structure can also help find related code when token
overlap is weak. AIRCoder~\citep{app_shi2026aircoder} preserves code structure during chunking.
Its retriever weights textual, dependency, and structural signals for each query.




%% file: sections/challenges.tex
\section{Open Challenges and Opportunities}
\label{sec:challenges}

RAG research has produced a family of methods for every stage of the lifecycle, and agentic
systems have moved the choice of workflow from the designer to the model at run time. Both leave
questions open, and this chapter discusses four of them: what a query is allowed to spend and who
decides that, what an agent is allowed to change, which step in the pipeline earned the credit, and
what the benchmarks actually measure.

\paragraph{Cost is per query.} Every stage has grown a family of methods of its own, and a
deployed system draws on several at once: a retriever, a reranker, a refiner, and a controller
now run in series. Each query pays for that complexity in tokens and latency, however much the
query actually needs. What keeps it affordable is a per-query control decision: how much of the
pipeline to run, and when. There are three ways to make it. It can be read off a signal the
model already produces, as FLARE \citep{flare} does when generation uncertainty rises; it can
come from a separate trained model, as Adaptive-RAG
\citep{adaptiverag} and RAGRouter \citep{ragrouter} do; or it can be asked of the LM itself, which then decides in a pass
of its own \citep{rowen,li2024longcontext}. Each way costs something different. A signal read
off the model is a proxy for knowing, and Section~\ref{sec:routing} notes how poorly models
judge what they know. A trained decision
needs labelled decisions and the compute to fit them, and it is fit to one model and one
corpus, so a change to either degrades it. A decision taken by the LM costs at least one extra
pass, and when the step it skips is a single lexical lookup, that pass costs more than the
lookup. The gap between regimes is large: Adaptive-RAG's three regimes cost $0.35$, $3.08$, and
$27.18$ seconds per query, and its trained router sends $9\%$, $53\%$, and $38\%$ of queries to
them \citep{adaptiverag}. Choosing wrongly is expensive: on the queries that need only a single
step, forcing the multi-step path costs nine times as much and returns $3.7$ F1 \emph{less}
\citep{adaptiverag}. What is missing is a decision that costs less than the step it skips and
is right more often than a fixed policy.

\paragraph{The index the agent cannot touch.} An agent chooses whether and what
to retrieve, but never what exists to be retrieved: in the systems we examined the
corpus is assembled before inference and is not modified by the agent. Search-R1 \citep{searchr1} trains its
policy against the 2018 Wikipedia dump with a single retriever; A-RAG \citep{arag} builds
a light hierarchy offline and notes that it avoids constructing graphs at all;
Graph-R1's reinforcement learning runs over a hypergraph built before training begins
\citep{graphr1}. The mirror image exists: AutoGraph-R1 \citep{autographr1} optimizes the constructor with
reward taken from a downstream RAG pipeline, while the retrieval server and answer generator
stay frozen. The two papers are the same loop cut at opposite ends, and
neither closes it: ``end-to-end'' in their titles is end-to-end over a trajectory or over a
construction, never over both. The cost is that the index is never a variable: every policy
the agent learns is optimal for the index it was given, and the global optimum over index and
policy is never sought. The next step is to put indexing under the agent's control.

\paragraph{Credit assignment for retrieval decisions.} Reinforcement learning made the
retrieval policy trainable, and reward is now the binding constraint. Search-R1 \citep{searchr1} optimizes a
simple outcome reward on the final answer, and Search-R2 \citep{searchr2} names the
consequence: a trajectory-level signal is sparse, fails to distinguish good reasoning from a
lucky guess, and induces credit misattribution across retrieval and reasoning steps. Step-level supervision exists at two prices, and neither buys a signal whose
validity is easy to check. It can be constructed offline, as DeepRAG \citep{deeprag} does by searching a binary
tree of retrieval decisions; the labels then cover the branches the search
visited, and a decision outside them gets none. It can be bought online, as HiPRAG \citep{hiprag} does by
regenerating each step's answer and asking an external judge whether the search was necessary,
with a cheaper non-reasoning judge to contain the cost; the label is now a second
model's opinion, and the cheaper judge is the less reliable one. The challenge is to get a reliable
step-level signal that is cheap and verifiable at once, which points at what retrieval already
returns \citep{igpo}.

\paragraph{Benchmarks that no longer match the task.} The agentic systems of
Section~\ref{sec:agentic} are evaluated almost exclusively on HotpotQA, MuSiQue,
2WikiMultihopQA, Natural Questions, TriviaQA, or PopQA. All six share three properties. The
evidence for a question is a sentence or two, so what they reward is locating and combining a
few passages rather than reaching knowledge the model lacks; that evidence also fits inside a
modern context window, and where the comparison has been run a long-context model wins by
$3.6$--$13.1\%$ on average, at a much higher price \citep{li2024longcontext}. The corpus is a
public dump older than the models now evaluated on it, so part of every score is recall rather
than retrieval. And the corpus
is fixed, so nothing tests a system that has to keep an index current. None of the three is the
case RAG is made for, which Section~\ref{sec:intro} states: knowledge the model was not trained
on, such as a client's files or yesterday's news. The benchmarks that do supply such corpora are
small. A benchmark with a private or growing corpus, evidence
that does not fit a single retrieval step, and leakage controlled by construction is missing,
and building one is a contribution in its own right.

%% file: sections/conclusion.tex
\section{Conclusion}
\label{sec:conclusion}

The lifecycle this review follows has five components: offline indexing, pre-retrieval, retrieval, post-retrieval, and orchestration, each with a family of methods of its own, and agentic systems change who decides the control flow. For each component the review moves from the problem, to the families of methods that address it, to the costs and limitations that remain. Evaluation and applications are discussed for better grounding. Four questions remain open: what a query is allowed to spend and who decides that, what an agent controls, which step earned the credit, and what the benchmarks measure. From tradition to present and beyond, this work provides a review of RAG from a perspective of the agentic era.

%% file: sections/ai.tex
\section*{Declaration on the Use of AI}
During the preparation of this manuscript, AI assistants based on generative language models were used to integrate manual notes and collected materials into draft prose, and to assist with language editing and proofreading, internal consistency checks of references and terminology, and the preparation of figures, tables, and the bibliography. The surveying itself was carried out entirely by the authors, who alone searched for and selected the literature, read and analyzed the surveyed papers, and formed the core viewpoints and organizing arguments of this review. All AI-assisted output was reviewed, verified, and edited by the authors, who take full responsibility for the content. No AI tool is listed as an author of this work. The use of AI does not impact the scientific rigor or originality of this work.